\documentclass[journal]{IEEEtran}
\usepackage{amsmath,amsfonts,amssymb,bm}
\usepackage{array}
\usepackage[table]{xcolor}
\usepackage[caption=false,font=normalsize,labelfont=sf,textfont=sf]{subfig}
\usepackage{textcomp}
\usepackage{stfloats}
\usepackage{flafter}
\usepackage{placeins}
\usepackage{url}
\usepackage{graphicx}
\usepackage{tikz}
\usetikzlibrary{calc}
\usepackage{float,multirow}
\usepackage{booktabs}
\usepackage{cite}
\usepackage{xr}[2024/04/10]

\newcommand{\modelname}{ExMR}

\tikzset{
    qualhighlight/.style={draw=orange!85!yellow,line width=0.65pt},
    qualinset/.style={draw=orange!85!yellow,line width=0.75pt},
    qualconnector/.style={draw=orange!85!yellow,line width=0.6pt,densely dashed}
}

\newlength{\qualbodyw}
\newlength{\qualcropw}
\newlength{\qualimgsep}
\newlength{\qualblocksep}
\newlength{\qualstacksep}

\definecolor{linkblue}{rgb}{0.16,0.32,0.62}
\usepackage[colorlinks,linkcolor=linkblue,citecolor=linkblue,urlcolor=linkblue,
    pdftitle={ExMR: Extremity Motion Refinement for Monocular Human Motion Capture},
    pdfauthor={Tom Wehrbein and Bodo Rosenhahn}]{hyperref}

\begin{document}

\title{ExMR: Extremity Motion Refinement for Monocular Human Motion Capture}

\author{Tom Wehrbein and Bodo Rosenhahn\\[2pt]
{\normalsize\url{https://twehrbein.github.io/exmr-website/}}
\thanks{This work has been submitted to the IEEE for possible publication. Copyright may be transferred without notice, after which this version may no longer be accessible.}%
\thanks{This work was supported by the Federal Ministry of Education and Research (BMBF), Germany, under the AI service center KISSKI (grant no.\ 01IS22093C), the Deutsche Forschungsgemeinschaft (DFG) under Germany's Excellence Strategy within the Cluster of Excellence PhoenixD (EXC 2122), and the European Union within the Horizon Europe research and innovation programme under grant agreement no.\ 101136006--XTREME.}%
\thanks{ Tom Wehrbein and Bodo Rosenhahn are with the Institute of Information Processing, Leibniz University Hannover, 30167 Hannover, Germany (e-mail: wehrbein@tnt.uni-hannover.de).}
}

\markboth{}%
{Wehrbein and Rosenhahn: \modelname: Extremity Motion Refinement for Monocular Human Motion Capture}

\maketitle

\begin{abstract}
Detailed hand and foot articulation is important for animation and motion analysis,
yet remains challenging to recover from monocular video.
Recent expressive video methods infer hand articulation from shared whole-body representations.
Our conference work suggests that this underexploits local evidence:
the Foot Motion Refinement (FootMR) module refines ankle rotations from foot keypoints,
improving reconstruction even when the initial estimator receives the same keypoints. 
We extend this principle to hands with the Hand Motion Refinement (HandMR) module.
Because 2D finger keypoints are often missing or unreliable, HandMR combines image features
of a pretrained hand reconstruction model with initial arm and hand rotations to
refine elbows, wrists, and fingers.
We integrate FootMR and HandMR into \modelname\ (Extremity Motion Refinement), a system for
reconstructing 3D body, hand, and foot motion from monocular video.
An initial temporal estimator predicts
whole-body motion, after which both refiners revisit local evidence to
predict residual joint rotations.
Experiments on ARCTIC, MOYO, RICH, and MOOF show that \modelname\ outperforms
state-of-the-art methods in hand and foot reconstruction, reducing
hand per-vertex error and ankle joint angle error on MOYO by
19\% and 18\%, respectively, relative to the strongest competitor.
Code will be made available for research purposes.

\end{abstract}


\begin{figure}[t]
    \centering
    \includegraphics{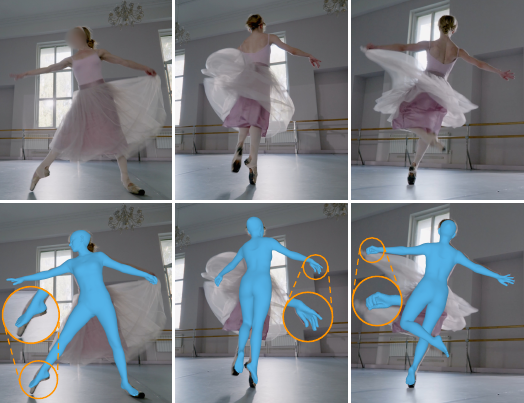}\par\prevdepth=0.0pt\relax
    \caption{
        \textbf{Monocular motion reconstruction with \modelname.}
        Our work focuses on the articulation of the extremities (the hands and feet)
        within full-body motion capture.
    }
    \label{fig:teaser}
\end{figure}

\section{Introduction}
\label{sec:intro}

\IEEEPARstart{M}{onocular} human motion capture aims to recover a person's 3D
movement from a single video, making motion reconstruction accessible for
applications such as animation, motion analysis, and AR/VR.
The motion of the extremities carries particularly important details:
ankle articulation contributes to lifelike dance animation, while wrist and finger
configurations convey how a hand grasps or manipulates an object.
Nevertheless, most current motion-capture methods primarily target overall body
motion~\cite{shin2024wham,wang2024tram,shen2024gvhmr,wang2025prompthmr,li2025genmo,wang2026duomo},
and recovering detailed hand and foot articulation remains challenging.

Single-image expressive estimators~\cite{cai2023smplerx,yin2025smplest,yang2026sam3dbody}
and per-frame body--hand integration methods~\cite{rong2021frankmocap,pymafx2023,moon2026h4wpp}
recover articulated hands with body context, but cannot exploit neighboring frames
to resolve pose ambiguities.
Hand-only video
methods~\cite{Duran_2024_WACV,zhang2025hawor,Ismayilzada_2026_CVPR} instead
use temporal evidence, but estimate the hand in isolation from the
articulated body that constrains it.
Expressive video methods such as WHAC~\cite{yin2024whac},
Motions as Queries~\cite{Liu_2025_CVPR}, and
Human3R~\cite{chen2026human3r} combine body and temporal evidence
to recover body motion and hand articulation jointly.
However, all three extract hand evidence from
person crops or full video frames, where the small image area of the hands
can limit the available visual detail~\cite{ExPose:2020}.
Closest to our work, concurrent DanceHMR~\cite{shen2026dancehmr} instead
fuses separately encoded body and specialist hand observations before a
shared temporal encoder.
After fusion, temporal modeling of body motion and hand articulation
relies exclusively on a shared whole-body representation, which receives the
hand observations only once, before an explicit estimate of the arms and hands
exists.

In our 3DV 2026 conference paper~\cite{wehrbein26footmr}, we introduced the
Foot Motion Refinement module (FootMR), which predicts residual ankle rotations
in a dedicated temporal network using local 2D foot keypoints, knee rotations,
and initial ankle rotations.
The paper showed that dedicated temporal refinement improved foot
reconstruction even when the initial temporal estimator already received the
same local evidence.
This finding motivates our hypothesis that fine-grained hand evidence
can remain underexploited in a shared whole-body representation.
We therefore investigate whether a
dedicated temporal stage can exploit the same hand observations more
effectively by revisiting them together with the arm and hand rotations of an
initial whole-body estimate and predicting residual joint rotations.
Learning corrections relative to an explicit estimate is
well established in single-image pose and mesh
recovery~\cite{carreira2016ief,kanazawa18hmr,ExPose:2020}.
Residual refinement has also been explored in video-based body reconstruction~\cite{luo2020meva},
including separate temporal networks for body regions in
Local Recurrent Models (LMR)~\cite{li2021localrecurrent}.
In our hand-refinement stage, the initial arm and hand rotations can
summarize broader body and temporal context absent from an individual hand
crop. Additionally, they can provide a useful prior when the local
evidence is ambiguous, for example, under occlusion.

Building on this refinement principle,
we introduce \modelname\ (\emph{Extremity Motion Refinement}), a system for
reconstructing body, hand, and foot motion from monocular video.
Its initial estimator extends the GVHMR-based~\cite{shen2024gvhmr} estimator
of our conference paper and fuses body and specialist hand
image features before temporal modeling to predict SMPL-X
motion~\cite{pavlakos2019smplx} with articulated hands.
FootMR and
the newly introduced Hand Motion Refinement module (HandMR) then refine the
initial motion sequence by predicting residual joint rotations in dedicated temporal networks.
We retain FootMR's architecture and refinement
formulation and train both refiners jointly with the initial estimator.
FootMR and HandMR instantiate a common temporal residual-refinement
principle: revisiting local evidence together with explicit initial joint rotations
to predict corrections over time.
Fig.~\ref{fig:teaser} shows \modelname's reconstructions of a dance
sequence, with close-ups highlighting the recovered foot and hand
articulation.

Adapting refinement to hands changes both the correction targets and the observations.
Hand refinement must account for a wrist and multiple finger joints per hand, compared with
one ankle rotation per side for foot refinement. A direct analog of FootMR would use
2D finger keypoints, but these are frequently missing or unreliable under self-occlusion
or occlusion by interacting objects. HandMR therefore uses latent
image features from HaMeR~\cite{pavlakos2024hamer}, a pretrained model for 3D
hand reconstruction.
Extracted from local hand crops, these features can retain
cues beyond explicit keypoint coordinates. HandMR reuses the HaMeR features supplied
to the initial estimator, combining them with initial arm--hand rotations to
jointly refine elbows, wrists, and fingers.

For feet, the design of FootMR was originally motivated by inaccurate foot
annotations and limited foot-motion diversity in the training data available
at the time of our conference work. Its keypoint representation permits
training on projected 3D motion without paired images and supports extensive
data augmentation. Advances in training data and image features could now
reduce the need for dedicated ankle refinement. BEDLAM2~\cite{tesch2025bedlam2} provides large-scale,
diverse synthetic video with ground-truth body motion, and
SAM3D-Body~\cite{yang2026sam3dbody} explicitly
targets accurate body, hand, and foot reconstruction. We therefore strengthen the
inherited body-and-foot system by adopting SAM3D-Body image features, revising its
training mixture and objectives, and updating its test-time keypoint detector.
Matched comparisons with and without FootMR test whether dedicated foot refinement
remains beneficial.

We evaluate hand and expressive whole-body reconstruction on
ARCTIC~\cite{fan2023arctic}, MOYO~\cite{tripathi2023moyo}, and
RICH~\cite{huang2022rich}, and foot reconstruction on MOYO, RICH, and the MOOF
dataset introduced in our conference paper~\cite{wehrbein26footmr}.
Among the evaluated competitors, \modelname\ achieves the
lowest hand reconstruction errors on all three hand benchmarks and
the best or tied-best results across the full-body metrics.
Compared with an otherwise identical variant trained
without HandMR, \modelname\ lowers hand errors on all three datasets, reducing
MOYO wrist-aligned hand per-vertex error from $39.9$ to $26.5\ \mathrm{mm}$.
Under the revised body-and-foot configuration,
adding the FootMR module improves foot reconstruction on all three foot
benchmarks, reducing MOYO ankle joint angle error from $25.3^\circ$ to
$20.4^\circ$ in a matched comparison.

Relative to our conference paper, this journal article makes three main
contributions:
\begin{itemize}
    \item We introduce HandMR within \modelname, extending our conference
    body-and-foot motion-capture system to articulated hands.
    HandMR follows FootMR's principle of predicting residual joint
  rotations in a dedicated temporal network. To adapt this principle to articulated
  hands, HandMR combines specialist hand image features with initial joint rotations
  to jointly correct elbow, wrist, and finger articulation.
    \item \modelname\ achieves the lowest hand reconstruction errors among
    the compared methods on all three hand benchmarks.
    HandMR substantially improves hand reconstruction over an otherwise
    identical \modelname\ variant trained without it.
    We further analyze hand-evidence representations, initial joint-rotation
    conditioning, temporal accuracy and stability, and how refinement gains
    vary with wrist articulation.
    \item \modelname\ outperforms all compared competitors on every
    reported foot metric across the three foot benchmarks.
    We show in matched comparisons that the inherited FootMR module
    improves foot reconstruction under the stronger body-and-foot configuration.
    A cumulative revision analysis traces the complete system's improvement
    over the conference configuration.
\end{itemize}

\section{Related Work}
\label{sec:related_work}

\subsection{Single-Image Human Mesh Recovery}

Single-image human mesh recovery commonly estimates the parameters of a
statistical body model from an RGB image. SMPL~\cite{loper2015smpl} provides
a compact representation of body pose and shape.
SMPLify~\cite{bogo16eccv} fits SMPL to detected 2D keypoints, while
HMR~\cite{kanazawa18hmr} learns feed-forward parameter regression.
A substantial literature has since developed around SMPL-based body
reconstruction.
CLIFF~\cite{li2022cliff} adds full-image bounding-box conditioning to account
for perspective effects, and HMR2.0~\cite{goel2023humans} improves visual representations with a large
vision transformer backbone.

SMPL-X~\cite{pavlakos2019smplx} extends SMPL with articulated hands and an
expressive face, enabling joint body, hand, and face recovery.
AiOS~\cite{sun2024aios} progressively refines body, hand, and face features
in a single network for expressive SMPL-X reconstruction.
SMPLer-X~\cite{cai2023smplerx} and SMPLest-X~\cite{yin2025smplest}
scale training data and transformer capacity for expressive reconstruction.
SAM3D-Body~\cite{yang2026sam3dbody} predicts Momentum Human Rig
(MHR)~\cite{ferguson2025mhr} parameters for body, hand, and foot recovery
using a DINOv3-based image backbone~\cite{simoni2026dinov}.
These single-image estimators cannot exploit neighboring frames to resolve
pose ambiguities and often produce temporal jitter.
Their learned visual representations nevertheless provide useful per-frame
evidence for \modelname's temporal reconstruction.

\subsection{Hand Reconstruction and Body--Hand Integration}

MANO~\cite{romero2017mano} is a parametric model of hand pose and shape
widely used for hand-only reconstruction.
HaMeR~\cite{pavlakos2024hamer} scales transformer capacity and hand training
data, and WiLoR~\cite{Potamias_2025_CVPR} combines hand localization with
MANO reconstruction and image-aligned refinement. These specialists provide
detailed local estimates, although self-occlusion and hand--object interaction
can leave their evidence incomplete.

Consistent body--hand reconstruction requires coordinating wrist orientation
with the arm pose while exploiting detailed local hand evidence.
One strategy reconciles separate body and hand predictions by adjusting the
arm pose during integration.
FrankMocap~\cite{rong2021frankmocap} compares direct copy-paste with
optimization-based fitting and learned integration that adjusts shoulder and
elbow rotations. PyMAF-X~\cite{pymafx2023} uses elbow-twist
compensation to avoid unnatural wrist poses.
Hand evidence can also guide body and wrist estimation directly:
PIXIE~\cite{feng2021pixie} fuses body and hand features, and
Hand4Whole~\cite{Moon_2022_CVPR} combines body and metacarpophalangeal (MCP)
joint cues for wrist estimation.
Such guidance can reuse a pretrained hand specialist, as in
Hand4Whole++~\cite{moon2026h4wpp}, which injects specialist features into a
frozen whole-body estimator to improve wrist orientation.
SAM3D-Body~\cite{yang2026sam3dbody} instead uses geometric prompts to refine
the body prediction before integrating optional hand-crop predictions.
In the converse direction, body estimates guide local hand reconstruction
through localization or explicit pose conditioning:
ExPose~\cite{ExPose:2020} uses an initial body estimate to define part crops,
whereas HMR-Adapter~\cite{shen2024hmradapter} conditions a frozen hand
estimator on joint rotations along the root-to-wrist kinematic chain.
These approaches establish complementary roles for local hand evidence and
body-derived kinematic context in recovering compatible arm and hand
configurations.

These integration methods operate per frame. Hand-only video methods exploit
temporal observations or motion priors to recover or refine hand motion,
without jointly estimating the articulated body.
HMP~\cite{Duran_2024_WACV} learns a hand-motion prior and
optimizes its latent variables to explain a video.
HaWoR~\cite{zhang2025hawor} estimates world-space hand trajectories from
egocentric video and infills motion during missing observations.
PAD-Hand~\cite{Ismayilzada_2026_CVPR}
uses physics-aware diffusion to refine initial hand pose sequences.
HandMR instead uses the arm--hand configuration from an initial whole-body
motion estimate to guide temporal hand refinement.

\subsection{Video-Based Human Motion Reconstruction}

Video-based methods exploit temporal context to reconstruct body
motion~\cite{kanazawa2019hmmr,kocabas2019vibe,luo2020meva,choi2020beyond,WeiLin2022mpsnet,shen2023global}.
Recent
methods~\cite{yuan2022glamr,ye2023slahmr,shin2024wham,wang2024tram,shen2024gvhmr,wang2025prompthmr,wang2026duomo,li2025genmo}
recover global motion in world coordinates while accounting for camera
motion, but do not estimate articulated finger motion.
Our initial estimator extends GVHMR's~\cite{shen2024gvhmr} temporal
transformer to articulated hands.

Expressive video methods jointly recover body motion and hand articulation,
with different sources of hand evidence.
Monocular Total Capture~\cite{xiang2019monocular} fits a unified model to
observations from separate body and hand networks, followed by temporal mesh
tracking.
Other approaches derive hand evidence from person crops or full video frames.
WHAC~\cite{yin2024whac} adds temporal modeling to
SMPLer-X~\cite{cai2023smplerx}, whose hand features are cropped from its
whole-body feature map.
Motions as Queries~\cite{Liu_2025_CVPR} and Human3R~\cite{chen2026human3r}
derive hand evidence from full-frame image features within their temporal
expressive estimators.
Concurrent DanceHMR~\cite{shen2026dancehmr} fuses separately encoded body and
specialist hand observations through a residual feature update before a
shared temporal encoder. Body and hand motion are decoded from its shared
representation.
The initial estimator of \modelname\ likewise fuses hand evidence before
temporal modeling. HandMR subsequently refines the predicted arm--hand motion
in a separate temporal stage.

\subsection{Foot Pose and Motion Reconstruction}

Ankle positions alone do not determine foot orientation.
The OpenPose foot dataset~\cite{cao2019openpose} and
COCO-WholeBody~\cite{jin2020whole} provide additional 2D annotations of the
big toe, small toe, and heel.
Whole-body 2D
detectors~\cite{xu2022vitpose+,jiang2024rtmw,khirodkar2024sapiens,khirodkar2026sapiens2}
estimate these landmarks together with the ankle, providing localized
observations for foot reconstruction.
Zhuo et al.~\cite{zhuo2023foot} use multiview data to reconstruct foot-pose
and contact supervision, then train a monocular body-and-foot regressor with
reversible kinematic topology to incorporate contact information.
SAM3D-Body~\cite{yang2026sam3dbody} explicitly targets foot pose within
single-image full-body reconstruction and reports a foot-specific evaluation
using six 2D foot keypoints.

Contact constraints, motion priors, and physics-based optimization also improve
reconstruction by reducing footskate and ground
penetration~\cite{zou2020footskate,rempe2020contact,shimada2020physcap,rempe2021humor}.
Such methods can update articulated pose, but their principal emphasis is
contact consistency and physical plausibility. Plausible ground contact alone
does not establish accurate ankle articulation.
Our conference paper~\cite{wehrbein26footmr} introduced the FootMR module,
which uses local 2D foot observations and temporal context to refine ankle
rotations in an initial body-motion estimate.
We retain FootMR as the foot-refinement component of \modelname.

\subsection{Human Pose and Motion Refinement}
\label{sec:related_refinement}

Single-image refinement methods use image evidence together with a current pose
or mesh estimate to guide corrections.
Iterative Error Feedback~\cite{carreira2016ief} predicts corrections to 2D
joint locations, whereas HMR~\cite{kanazawa18hmr} iteratively updates body pose,
shape, and camera parameters.
PyMAF~\cite{zhang21pymaf} and ReFit~\cite{wang2023refit} use the predicted
geometry to query image features and iteratively improve alignment.
Other approaches refine predicted 2D
poses~\cite{fieraru2018refine,moon2019posefix} or optimize an initial
mesh using dense image correspondences~\cite{wehrbein2023iccvw}.

Several temporal refiners operate on pose-sequence representations alone.
PoseBERT~\cite{baradel2023posebert} is trained by masked modeling on
motion-capture data and refines both SMPL and MANO sequences.
SmoothNet~\cite{zeng2022smoothnet} refines coordinate- or rotation-based pose
sequences by modeling each component independently over time.
SynSP~\cite{wang2024synsp} uses pose-quality cues to improve the accuracy
and temporal smoothness of predicted pose sequences.
These methods are trained independently of the upstream pose estimators whose
outputs they correct.

ExPose~\cite{ExPose:2020} and MEVA~\cite{luo2020meva} integrate
residual correction of an initial estimate into image- and video-based
reconstruction, respectively.
ExPose combines local hand crops with the body estimator's wrist and finger
parameters in a single-image hand refiner, alongside a separate face refiner.
MEVA first estimates coarse body motion by mapping video features into the
latent space of a motion variational autoencoder.
A residual MLP then refines whole-body pose and shape per frame using
current estimates and temporally encoded visual features.
Temporal dynamics can also be modeled separately for body
regions~\cite{li2021localrecurrent,lee2021partdynamics}.
Local Recurrent Models (LMR)~\cite{li2021localrecurrent} use regional
temporal modeling to predict residual updates:
six recurrent networks iteratively update body regions from mean SMPL pose and
shape, each receiving the same per-frame image features together with its
current regional pose and the current shape estimate.
Compared with ExPose, HandMR additionally conditions on and refines elbow
rotations and uses temporal context.
MEVA and LMR use body-level image features, whereas FootMR and HandMR
receive observations specific to their target regions: 2D foot keypoints
and specialist hand image features, respectively.
Like LMR, our refiners use separate temporal networks with explicit pose
conditioning. In \modelname, these networks revisit the local foot and hand
observations,
together with its predicted joint rotations, to refine only selected joints.

\section{Method}
\label{sec:method}

\subsection{Problem Formulation and System Overview}
\label{sec:system_overview}

\begin{figure*}[t]
    \centering
    \includegraphics[width=\textwidth]{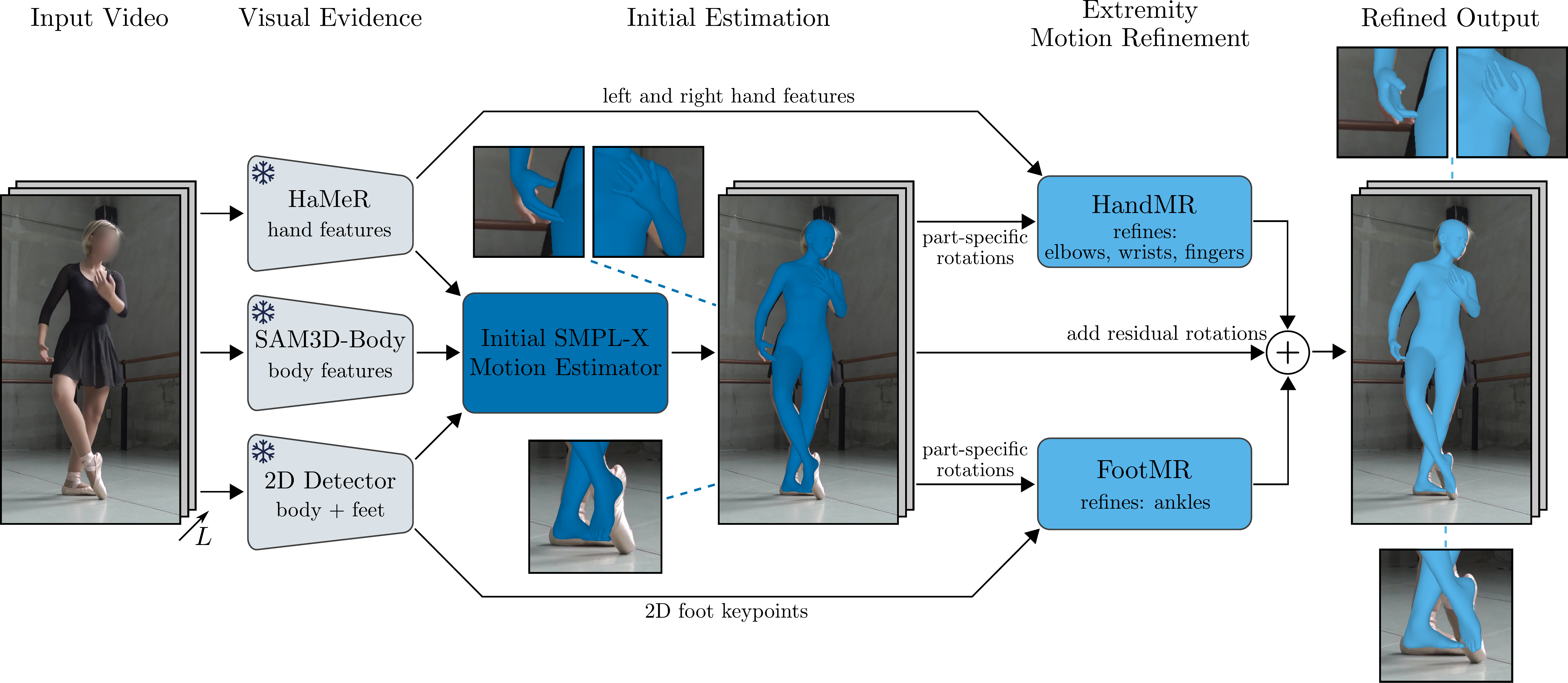}
    \caption{
        \textbf{Overview of \modelname\ at inference.}
        Given a monocular video, the initial estimator combines per-frame body and hand
        image features with 2D body-and-foot keypoints to predict a SMPL-X
        motion sequence. The HandMR and FootMR modules reuse the corresponding
        visual evidence and predict residual rotations conditioned on part-specific
        initial rotations: elbows, wrists, and fingers for HandMR, and knees and
        ankles for FootMR. HandMR refines elbows, wrists, and fingers, while FootMR
        refines ankles. The circled plus denotes addition of 6D rotation residuals,
        with conversion to parent-relative SMPL-X rotations where needed. The initial estimator
        and both refinement modules are jointly trained temporal transformers.
        Snowflakes denote frozen pretrained models. Additional inputs are
        omitted for clarity: the person bounding-box encoding for all three
        temporal networks and the relative camera rotation for the initial estimator.
        Insets compare the initial and refined hand and foot estimates.
    }\label{fig:method}
\end{figure*}

Given a monocular video of $L$ frames, we recover the person's
SMPL-X~\cite{pavlakos2019smplx} motion, with particular emphasis
on the articulated feet and hands. In the camera frame, we parameterize the
motion by joint rotations
$\{\theta^{(l)}\in\mathbb{R}^{52\times 6}\}_{l=1}^{L}$, shape
$\beta\in\mathbb{R}^{10}$, and camera-frame translations
$\{\tau_c^{(l)}\in\mathbb{R}^{3}\}_{l=1}^{L}$. Joint rotations use the
continuous 6D representation~\cite{zhou19rot6d}. Facial pose and expression
are outside the scope of this work.

As shown in Fig.~\ref{fig:method}, our system follows a two-stage design. Our
initial estimator, which builds on GVHMR~\cite{shen2024gvhmr}
(Section~\ref{sec:initial_estimator}), first predicts an initial full-body sequence
with articulated hands. Two temporal refinement modules then update its
extremity rotations: the \emph{Foot Motion Refinement} module (FootMR,
Section~\ref{sec:foot_refinement}), inherited from our conference
paper~\cite{wehrbein26footmr}, refines the two ankles, while the newly introduced
\emph{Hand Motion Refinement} module (HandMR,
Section~\ref{sec:hand_refinement}) refines both elbows, both wrists, and all 30
finger joints. Both modules predict residual 6D
rotations, which we fuse with the initial estimate to obtain the final
camera-frame sequence. We call the resulting system \modelname\
(\emph{Extremity Motion Refinement}).
Throughout the paper, FootMR and HandMR, when used on their own, denote the two
refinement modules, and \modelname\ denotes the complete reconstruction system.

The two refiners use modality-specific observations: FootMR receives sparse 2D
foot keypoints, whereas HandMR receives
image features from a pretrained 3D hand reconstruction model.
Section~\ref{sec:refiner_architecture} describes the refiners' temporal
architectures, and Section~\ref{sec:training} presents the objectives and
supervision used to train the complete system jointly.

\subsection{Initial Temporal SMPL-X Estimator}
\label{sec:initial_estimator}

Following the system design introduced in our conference
paper~\cite{wehrbein26footmr}, we build the
initial estimator as a temporal transformer based on
GVHMR~\cite{shen2024gvhmr} and extend it to produce full-body SMPL-X motion
with articulated hands.
The temporal transformer uses 12 encoder layers with 512-dimensional tokens.
In this work, only its camera-frame predictions enter the extremity refiners
and quantitative evaluations. We retain GVHMR's world-grounded recovery path
for demonstrations but do not evaluate it as a contribution.

For each frame, the initial estimator receives GVHMR's camera-motion
conditioning (a 6D
relative camera rotation), 23 normalized 2D body-and-foot keypoints, a
CLIFF-style three-value encoding of the person bounding
box, one 1024-dimensional SAM3D-Body image
feature~\cite{yang2026sam3dbody}, and one
1024-dimensional HaMeR feature~\cite{pavlakos2024hamer} for each hand,
extracted from a separate per-hand image crop.
The initial estimator uses the 23-keypoint layout from our conference
paper~\cite{wehrbein26footmr}. The layout comprises the 17 COCO~\cite{lin2014coco} body
keypoints used by GVHMR and three standard whole-body detector outputs per foot:
the big toe, small toe, and heel. Each keypoint is normalized by
subtracting the bounding-box center and dividing by its scale.
Keypoints deemed not visible according to their detector confidence are set to
zero.
Following CLIFF~\cite{li2022cliff}, the bounding-box encoding contains the person-box
center offsets and scale in the full image, normalized by the camera focal
length. It therefore preserves the location and apparent size removed by
keypoint normalization, which are informative under perspective projection.

Relative to GVHMR, we replace its HMR2.0 image feature~\cite{goel2023humans}
with the SAM3D-Body image feature, add the two HaMeR features, and extend the
output with 15 finger-joint rotations per hand. A learned projection shared
between hands embeds the added hand features. The resulting embeddings are
concatenated, then fused with the remaining input embeddings by element-wise
summation to form the per-frame temporal token. The initial
estimator predicts the joint rotations and shape together with
weak-perspective camera parameters for each person crop. The
sequence-level shape $\beta$ is obtained by averaging the predicted shape coefficients
over all $L$ frames. Following
CLIFF, we convert the camera parameters to the
perspective-camera translation $\tau_c^{(l)}$ for the full image, thereby
obtaining the initial camera-frame SMPL-X sequence. FootMR and HandMR
subsequently update only their designated joint rotations. Neither the shape
estimate nor the camera estimate changes during refinement.

\subsection{Dedicated Part-Specific Temporal Refinement}
\label{sec:part_specific_refinement}

FootMR and HandMR predict residual updates to different extremity joint
groups from sequences of local observations and initial joint rotations.
We first establish the rotation notation and update operation shared by the
two refiners.
SMPL-X expresses each joint rotation relative to its parent. Let
$\operatorname{rotmat}:\mathbb{R}^6\to\mathrm{SO}(3)$ map a 6D rotation
representation to its rotation matrix via Gram--Schmidt
orthonormalization~\cite{zhou19rot6d}, and let
$\operatorname{rot6d}:\mathrm{SO}(3)\to\mathbb{R}^6$ map a rotation matrix back
to its 6D representation. We write $\theta_i^{(l)}\in\mathbb{R}^6$ for the
parent-relative rotation of joint $i$ in frame $l$ and
$\mathbf{R}_i^{(l)}=\operatorname{rotmat}(\theta_i^{(l)})$ for its rotation
matrix. Its global rotation in the camera frame is
\begin{equation}
    \bar{\mathbf{R}}_i^{(l)}=
    \left(\prod_{j\in\mathcal{A}(i)}^{\rightarrow}
    \mathbf{R}_j^{(l)}\right)\mathbf{R}_i^{(l)},
\end{equation}
where $\prod^{\rightarrow}$ denotes ordered matrix multiplication along the
kinematic chain from the root toward $i$, and $\mathcal{A}(i)$ contains the
ancestors of $i$ in that order. We
write $\bar{\theta}_i^{(l)}=\operatorname{rot6d}(\bar{\mathbf{R}}_i^{(l)})$ for the resulting global rotation of joint $i$ in 6D representation.

Each refiner predicts a 6D residual vector for every target joint and frame.
Each residual is added to the corresponding initial 6D rotation in the
representation chosen for that joint, either global or parent-relative.
The updated 6D vector is mapped to a rotation matrix with
$\operatorname{rotmat}$. Rotations refined globally are then converted back
to the parent-relative representation required by SMPL-X. The following
subsections specify the observations, conditioning rotations, and correction
targets of each refiner.

\subsubsection{Foot Motion Refinement}
\label{sec:foot_refinement}

FootMR jointly refines the initial left and right ankle rotations using sparse
2D foot evidence. Although SMPL-X parameterizes each foot with an ankle and a
forefoot joint, we target only the ankle. It has the larger influence on
overall foot motion, and our prior study found only limited gains from
additionally refining the forefoot~\cite{wehrbein26footmr}. Four 2D keypoints
per foot---the ankle, big toe, small toe, and
heel---provide compact, direct geometric evidence for this target and can be
reliably detected by modern whole-body pose
estimators~\cite{khirodkar2026sapiens2}. Moreover, to our knowledge, no
publicly available foot-specific monocular 3D pose model exposes image features
analogous to HaMeR's hand features.
For each frame, FootMR receives these four 2D keypoints per foot, the
person bounding-box encoding, and the global rotations of both knees and
ankles. Only the ankle rotations are refined; the knee rotations provide
kinematic context. As in the initial estimator, keypoints deemed not visible
according to their detector confidence are set to zero after normalization.

A hip-relative knee rotation alone does not specify the knee orientation in
the camera frame because it omits the rotations of its ancestors. We therefore
condition FootMR on global knee rotations, which constrain feasible global
ankle rotations and help resolve the ambiguity in lifting the 2D keypoint
sequence to 3D.
The initial global ankle rotations provide a strong prior, particularly when
the 2D observations are noisy or missing. Using global rather than
knee-relative initial ankle rotations as the reference for residual refinement
also improves generalization to extreme ankle motions.

Let $a$ denote the ordered left--right ankle pair. FootMR predicts a pair of 6D
residuals $\delta\bar{\theta}_a^{(l)}\in\mathbb{R}^{2\times6}$, one for each
ankle. The refined global rotations are
\begin{equation}
    \bar{\theta}_a^{(l)}=
    \bar{\theta}_{a,\mathrm{init}}^{(l)}+
    \delta\bar{\theta}_a^{(l)}.
\end{equation}
Each updated 6D vector is mapped to a rotation matrix as
$\bar{\mathbf{R}}_a^{(l)}=
\operatorname{rotmat}(\bar{\theta}_a^{(l)})$.
We then recover the knee-relative ankle rotations independently for each
side as
\begin{equation}
    \mathbf{R}_a^{(l)}=
    \left(\bar{\mathbf{R}}_k^{(l)}\right)^{\!\top}
    \bar{\mathbf{R}}_a^{(l)},
\end{equation}
where $k$ denotes the knee on the same side.
These recovered knee-relative rotations replace the initial ankle rotations
when constructing the output SMPL-X sequence.

During training, FootMR conditions on ground-truth global knee rotations and
initial global ankle rotations predicted by the initial estimator. At
inference, both the knee and ankle conditions come from the initial estimator.
These rotation conditions are constructed without gradient tracking.

\subsubsection{Hand Motion Refinement}
\label{sec:hand_refinement}

HandMR specializes the refinement principle for articulated hands. A direct
analog of FootMR would use detected 2D finger keypoints for each hand.
Although these keypoints provide detailed geometric evidence when visible,
parts of the fingers are frequently self-occluded or hidden by objects or the
other hand, making the corresponding detections missing or unreliable. We
therefore use image features from HaMeR~\cite{pavlakos2024hamer}, a specialist
model trained on a large mixture of 2D- and 3D-annotated hand data and designed
for robust monocular 3D hand reconstruction.
We choose HaMeR because its transformer decoder provides a single
1024-dimensional output token per hand crop, which can be used directly
as input to our temporal networks. Compared with a coordinate-only
representation, its latent features retain richer image evidence and can
preserve cues about visibility and ambiguity under partial occlusion. Our ablation in
Table~\ref{table:hand_integration_evidence_ablation} shows that these features provide more effective
hand evidence than 2D finger keypoints in our complete pipeline.

HaMeR predicts a global hand-root orientation in its hand-crop camera frame and
parent-relative finger rotations. HandMR similarly refines wrists
globally in the full-image camera frame and fingers relative to their parents.
It additionally conditions on and refines the global elbow rotations. Since
HandMR predicts each wrist rotation globally whereas SMPL-X stores it relative
to the elbow, the accuracy of the recovered elbow-relative wrist rotation
depends on both the global wrist and elbow estimates. Refining the elbow, rather
than using it only as context, improves the consistency of the elbow--wrist
kinematic chain (Table~\ref{table:elbow_refinement}).

HandMR jointly
processes the HaMeR feature vectors, the person
bounding-box encoding, and the initial estimator's predictions for both arms:
the global rotations of both elbows and wrists and the parent-relative
rotations of all 30 finger joints. It jointly predicts residual rotations for
the left and right elbows, wrists, and fingers. Let $e$, $w$, and $f$ denote the
elbow, wrist, and finger joint groups, respectively. The residuals then have
dimensions $\delta\bar{\theta}_e^{(l)}\in\mathbb{R}^{2\times6}$,
$\delta\bar{\theta}_w^{(l)}\in\mathbb{R}^{2\times6}$, and
$\delta\theta_f^{(l)}\in\mathbb{R}^{30\times6}$. The updates are
\begin{align}
    \bar{\theta}_e^{(l)}
        &=\bar{\theta}_{e,\mathrm{init}}^{(l)}
          +\delta\bar{\theta}_e^{(l)}, \\
    \bar{\theta}_w^{(l)}
        &=\bar{\theta}_{w,\mathrm{init}}^{(l)}
          +\delta\bar{\theta}_w^{(l)}, \\
    \theta_f^{(l)}
        &=\theta_{f,\mathrm{init}}^{(l)}+\delta\theta_f^{(l)}.
\end{align}
The elbow and wrist residuals operate in the global camera frame, whereas the
finger residuals operate in the parent-relative representation.

Independently for each side, we convert the refined elbow and wrist rotations
back to the SMPL-X kinematic representation. First, each updated 6D vector is
mapped to a matrix,
$\bar{\mathbf{R}}_e^{(l)}=
\operatorname{rotmat}(\bar{\theta}_e^{(l)})$ and
$\bar{\mathbf{R}}_w^{(l)}=
\operatorname{rotmat}(\bar{\theta}_w^{(l)})$.
The parent-relative rotations are then recovered as
\begin{equation}
    \mathbf{R}_e^{(l)}=
    \left(\bar{\mathbf{R}}_s^{(l)}\right)^{\!\top}
    \bar{\mathbf{R}}_e^{(l)},
    \qquad
    \mathbf{R}_w^{(l)}=
    \left(\bar{\mathbf{R}}_e^{(l)}\right)^{\!\top}
    \bar{\mathbf{R}}_w^{(l)},
\end{equation}
where $s$ denotes the shoulder on the same side and
$\bar{\mathbf{R}}_s^{(l)}$ is its unrefined global rotation from the initial
estimate. Importantly, the second conversion uses the refined elbow rotation.

The predicted elbow, wrist, and finger conditions remain connected to the
initial estimator's computational graph, allowing gradients from HandMR to
propagate into the initial estimator. In preliminary
experiments, this configuration gave slightly better reconstruction accuracy
than detaching the HandMR conditions, so we use it in the final configuration.
The initial estimator and HandMR receive the same HaMeR features and
person bounding-box encoding. HandMR thus introduces a dedicated temporal
refinement stage rather than additional external hand evidence.

\subsubsection{Temporal Refiner Architectures}
\label{sec:refiner_architecture}

FootMR and HandMR are separate temporal networks that update disjoint joint
groups without sharing parameters. Each independently processes its own
evidence and part-specific predictions from the shared initial estimator.
Both networks instantiate the same
architectural blueprint, which we adopt from GVHMR~\cite{shen2024gvhmr}.

Following GVHMR, dedicated MLPs first embed the part-specific evidence, the
person bounding-box encoding, and rotation conditions. FootMR combines its
embeddings through element-wise summation into a 256-dimensional token. For
HandMR, a shared learned projection independently maps each of the two
1024-dimensional HaMeR
feature vectors to 256 dimensions. The resulting embeddings are concatenated
into a 512-dimensional visual-evidence embedding and summed element-wise with
the 512-dimensional box and rotation embeddings to form a 512-dimensional
token. This larger token accommodates HandMR's higher-dimensional hand evidence
and substantially larger set of conditioned rotations compared with FootMR's
few 2D keypoints and joint rotations. In either network, all information for
both sides of the body is represented by a single token per frame. The token
sequence is processed by six transformer encoder layers with four attention
heads and Rotary Position Embedding (RoPE)~\cite{su2024rope}. A two-layer MLP
with GELU activation maps each output token to all residual rotations for the
joint groups refined by the corresponding network.

RoPE encodes relative temporal position, while a local attention mask restricts
each token to a $W$-frame neighborhood. We use $W=120$, corresponding to four
seconds at 30 frames per second. This combination allows the refiners to
process sequences of arbitrary length in a single forward pass, without
autoregressive or sliding-window inference. The initial estimator uses the same
relative-position and local-attention principle inherited from GVHMR, so the
complete temporal model supports this sequence-level inference.

The initial estimator, FootMR, and HandMR contain $41.4$M, $3.6$M, and
$13.9$M trainable parameters, respectively, totaling $58.9$M for
\modelname, excluding the frozen feature extractors. FootMR and
HandMR have low absolute inference runtimes. Their individual contributions to
the complete system's cost are reported in the supplementary material
(Section~\ref{sec:inference_time_analysis}).

\subsection{Training Objectives and Supervision}
\label{sec:training}

The initial estimator, FootMR, and HandMR are trained jointly from scratch.
The pretrained SAM3D-Body and HaMeR feature extractors remain fixed, and their
output features are precomputed. During training, we synthesize the 23 input
keypoints by extracting 3D joints from the ground-truth SMPL-X sequence,
perturbing them with noise, and projecting them to 2D,
following prior temporal motion reconstruction
methods~\cite{shin2024wham,shen2024gvhmr,shin2023bio}.
Following GVHMR, random visibility masks simulate missing
detector outputs in these synthetic inputs.

We retain the random global-orientation augmentation from the FootMR module
introduced in our conference paper~\cite{wehrbein26footmr}, which is enabled by
the module's
image-independent evidence. For each sequence, we sample a
random rotation $\mathbf{R}_{\mathrm{aug}}\sim\mathcal{U}(\mathrm{SO}(3))$ and apply it to the
ground-truth and predicted root orientations before constructing FootMR's 2D
foot keypoints, bounding boxes, and global knee and ankle conditions. This
exposes the refiner to the same relative foot motion under arbitrary global
orientations and improves generalization to uncommon global ankle orientations.
No output adjustment is required because the refined global ankle rotations are
converted back to knee-relative rotations, which are independent of the root
orientation.
This augmentation is applied only when constructing FootMR's inputs, not when
constructing the inputs of the initial estimator or HandMR.
The precomputed SAM3D-Body and HaMeR features used by these components
encode appearance in the original image and cannot be transformed consistently
under a sampled virtual orientation.

We integrate the refined rotations into the SMPL-X prediction. The refiners
require no separate objectives. Let
$\Theta_{\mathrm{init}}$, $\Theta_{\mathrm{ref}}$,
and $\Theta^{*}$ denote the initial, refined, and
ground-truth SMPL-X rotation sequences in $\mathbb{R}^{L\times52\times6}$,
obtained by stacking the per-frame 6D rotations $\theta^{(l)}$. The initial and
refined sequences differ only in the rotations updated by the two refiners:
both ankles, both elbows, both wrists, and all 30 finger joints. Both estimates
are supervised with
\begin{equation}
    \mathcal{L}_{\theta}=\tfrac{1}{2}\left(
    \ell_2(\Theta_{\mathrm{init}},\Theta^{*})+
    \ell_2(\Theta_{\mathrm{ref}},\Theta^{*})\right).
\end{equation}
Throughout this
subsection, $\ell_1(\cdot,\cdot)$ and $\ell_2(\cdot,\cdot)$ denote losses based
on element-wise absolute and squared errors, respectively.
Body-shape supervision and the inherited world-motion objectives remain
unchanged from GVHMR~\cite{shen2024gvhmr}.

Camera-frame geometric losses additionally supervise the refined prediction.
Let $\mathbf{J}\in\mathbb{R}^{L\times63\times3}$ and
$\mathbf{V}\in\mathbb{R}^{L\times437\times3}$ denote its SMPL-X joint and
supervised-vertex sequences, with starred symbols denoting ground truth.
Following GVHMR, the 437 vertices are sampled across the full SMPL-X mesh of
$10{,}475$ vertices. The first 23 entries along the joint dimension of
$\mathbf{J}$ are body-and-foot joints, with index set $\mathcal{B}$, while the
remaining 40 are finger joints and fingertips, with index set $\mathcal{H}$.
Their root-centered versions, $\mathbf{J}^{r}$ and $\mathbf{V}^{r}$,
are formed by subtracting the hip midpoint in each frame. For
$\mathbf{J}_{\mathcal{H}}$, we also form per-hand wrist-centered coordinates,
denoted by $\mathbf{J}^{\mathrm{wr}}_{\mathcal{H}}$, by centering each hand at its
corresponding wrist.
We retain the forms of GVHMR's camera-frame 3D body-joint and vertex losses and
its 2D joint and vertex reprojection losses, but replace their $\ell_2$
penalties with $\ell_1$. The 2D joint term is extended to the full 63-joint set,
and a wrist-centered 3D finger-joint term additionally supervises articulated
hands.
The resulting 3D and 2D losses are
\begin{align}
    \mathcal{L}_{\mathrm{3d}}={}&
        \ell_1(\mathbf{J}^{r}_{\mathcal{B}},
        \mathbf{J}^{r*}_{\mathcal{B}})
        +\ell_1(\mathbf{J}^{\mathrm{wr}}_{\mathcal{H}},
        \mathbf{J}^{\mathrm{wr}*}_{\mathcal{H}})\notag\\
        &+\ell_1(\mathbf{V}^{r},\mathbf{V}^{r*}),\\
    \mathcal{L}_{\mathrm{2d}}={}&
        \ell_1(\Pi(\mathbf{J}),\Pi(\mathbf{J}^{*}))
        +\ell_1(\Pi(\mathbf{V}),\Pi(\mathbf{V}^{*})),
\end{align}
where $\Pi$ projects joints and vertices framewise into normalized bounding-box
coordinates. The crop-camera supervision likewise follows GVHMR, with an $\ell_1$
penalty replacing the original $\ell_2$ penalty. For the GVHMR-derived camera-frame geometric terms,
replacing $\ell_2$ with $\ell_1$ changes their numerical scale, so we rescale
their weights accordingly.

We introduce $\ell_1$ losses on the first- and second-order frame differences of
the root-centered 63-joint sequence, which we refer to as velocity and
acceleration losses. For frame index $l$, let
$\Delta\mathbf{J}^{r,(l)}=\mathbf{J}^{r,(l+1)}-\mathbf{J}^{r,(l)}$ and
$\Delta^2\mathbf{J}^{r,(l)}=\mathbf{J}^{r,(l+2)}
-2\mathbf{J}^{r,(l+1)}+\mathbf{J}^{r,(l)}$, for $l=1,\ldots,L-1$ and
$l=1,\ldots,L-2$, respectively. The temporal losses are
\begin{align}
    \mathcal{L}_{\mathrm{vel}}
        &=\ell_1(\Delta\mathbf{J}^{r},
          \Delta\mathbf{J}^{r*}),\\
    \mathcal{L}_{\mathrm{acc}}
        &=\ell_1(\Delta^2\mathbf{J}^{r},
          \Delta^2\mathbf{J}^{r*}).
\end{align}
Temporal differences enter the losses only when all contributing frames are
valid.

Unavailable supervision is masked. In particular, H36M~\cite{ionescu2014h36m}
and 3DPW~\cite{marcard2018eccv} (two
datasets in the training mixture detailed in
Section~\ref{sec:experimental_protocol}) lack articulated hand annotations.
Their elbow, wrist, and finger rotations, finger joints, hand-driven vertices,
and corresponding temporal terms are therefore masked.

\noindent\textbf{Data provenance.}
This study uses existing published datasets, including MOOF introduced in our
prior work~\cite{wehrbein26footmr}, and publicly available Pexels videos\footnote{\url{https://www.pexels.com/videos/}}
for qualitative evaluation. No new participants were recruited, and no new
recordings of participants were collected for this study.

\section{Experiments}
\label{sec:experiments}

We first describe the experimental protocol and evaluate expressive whole-body
and hand reconstruction. The subsequent analyses examine HandMR's
contribution and behavior and revisit foot refinement under the revised
configuration, including qualitative examples of hand and foot refinement.
A secondary body-only comparison concludes the evaluation.

\subsection{Experimental Protocol}
\label{sec:experimental_protocol}

\subsubsection{Evaluation Datasets}
We evaluate expressive whole-body and hand reconstruction on
ARCTIC~\cite{fan2023arctic}, MOYO~\cite{tripathi2023moyo}, and
RICH~\cite{huang2022rich}, which provide SMPL-X ground truth with
articulated hand pose. ARCTIC records dexterous bimanual human--object
interaction, whereas MOYO captures a yoga professional performing
complex poses. Both pair multiview video with marker-based motion capture
to obtain the annotations. RICH records multiple
subjects performing daily and sporting activities indoors and outdoors, with
annotations obtained through markerless multiview capture.

On all three datasets, we evaluate every camera view independently as a
monocular sequence. Because ARCTIC's test-set annotations are withheld, we
follow its official split, evaluating on the validation subject and restricting
ARCTIC training to the eight training subjects.
MOYO separates sequences captured with and without hand markers. We therefore
compute metrics requiring articulated hand ground truth only on the
release-defined \texttt{with\_hands} validation subset. Full-body PVE and
PA-PVE, as well as prediction-only hand jitter, use the complete validation set.

For fine-grained foot evaluation, we use MOYO and RICH for 3D reconstruction
and MOOF for 2D reprojection. Introduced in our conference
paper~\cite{wehrbein26footmr}, MOOF contains complex foot motion with manually
verified 2D foot-keypoint and visibility annotations.
Additionally, body-only reconstruction is reported on 3DPW~\cite{marcard2018eccv},
EMDB-1~\cite{kaufmann2023emdb}, and RICH.

\subsubsection{Evaluation Metrics}
We evaluate 3D joint and vertex errors in camera coordinates. Whole-body
mean per-vertex error (PVE) uses all SMPL-X vertices, whereas hand PVE uses
MANO-corresponding vertices. Body-only evaluation uses SMPL meshes to compute
PVE and mean per-joint position error (MPJPE).
Body metrics are pelvis-aligned and hand metrics are
wrist-aligned.
The Procrustes-aligned variants (PA-PVE and PA-MPJPE) additionally apply
per-frame similarity alignment, independently for each hand in hand PA-PVE.
These errors are reported in millimeters.

Before SMPL-X mesh construction, we set facial pose and expression to
neutral defaults for all locally evaluated methods and ground truth.
For body-only evaluation, we also set articulated hand poses to their
defaults in both predictions and ground truth. SMPL-X meshes are then mapped
to the $6{,}890$-vertex SMPL topology~\cite{pavlakos2019smplx} before
computing PVE and regressing joints for MPJPE and PA-MPJPE.

Temporal metrics use joint trajectories at $30\ \mathrm{frames/s}$, centered in each frame.
Body-only Accel uses the pelvis-centered MPJPE joint set, and hand metrics use
21 wrist-centered joints per hand, including fingertips. Finite differences
are scaled by the frame rate raised to their order. Mean per-joint velocity
error (MPJVE, $\mathrm{mm/s}$) and acceleration error (Accel,
$\mathrm{m/s^2}$) measure mean Euclidean discrepancies between predicted and
ground-truth first- and second-order differences, respectively.
Prediction-only Jitter measures the mean magnitude of third-order
differences, reported in units of $10\ \mathrm{m/s^3}$ following
WHAM~\cite{shin2024wham}.

Global and local rotation errors (G-RE and L-RE) measure the $\mathrm{SO}(3)$
geodesic distance between predicted and ground-truth rotations, in degrees.
G-RE uses global camera-frame rotations, whereas L-RE uses parent-relative
rotations. We report ankle-specific G-RE as global ankle joint angle error
(AJAE).

Whole-body averages can obscure localized foot improvements~\cite{wehrbein26footmr}.
We therefore also report scale-normalized foot MPJPE
(N-MPJPE\textsubscript{F}, in millimeters) over each foot's big toe, small toe,
and heel. For each foot and frame,
predicted and ground-truth landmarks are mean-centered and then
scale aligned~\cite{rhodin2018cvpr}.
On MOOF, the percentage of correct foot keypoints
(PCK\textsubscript{F}) pools visible landmarks using an $\ell_2$
reprojection-error threshold of $0.05$ times the person box's larger side.
Because PCK\textsubscript{F} also depends on body pose, shape, and translation,
we report normalized 2D foot-keypoint error (N-FKE\textsubscript{2d}) to
isolate local foot configuration. After person-box normalization, each foot's
keypoints are mean-centered and scale-aligned independently per frame before
their $\ell_2$ error is computed.
N-FKE\textsubscript{2d} includes only feet with all three landmarks visible.

\subsubsection{Training and Implementation Details}
The training set comprises BEDLAM~\cite{black2023bedlam},
BEDLAM2~\cite{tesch2025bedlam2}, H36M~\cite{ionescu2014h36m},
3DPW~\cite{marcard2018eccv}, and ARCTIC~\cite{fan2023arctic}. Compared with the
conference training mixture~\cite{wehrbein26footmr}, we add BEDLAM2 and ARCTIC
and remove AMASS~\cite{mahmood2019amass} because it lacks images for HaMeR feature
extraction.
To prevent overlap with the MOYO evaluation set, BEDLAM2 sequences
whose identifiers contain \texttt{moyo} are excluded. The five datasets are concatenated
and globally shuffled without rebalancing.
\modelname\ predicts neutral SMPL-X parameters.
For H36M and 3DPW, we use the fits from GVHMR~\cite{shen2024gvhmr}, whereas
BEDLAM and BEDLAM2 directly provide ground-truth parameters.
For ARCTIC, we fit shape coefficients to the personalized
ground-truth meshes of the training subjects
(Section~\ref{sec:arctic_shape_fitting}).

Training uses dataset-provided ground-truth camera intrinsics and person
bounding boxes. Each hand box is constructed by projecting the annotated
SMPL-X hand joints. We cap training subsequences at 120 frames and use a batch
size of 128, which keeps training of the complete \modelname\ system within
$24\ \mathrm{GB}$ of GPU memory.
The camera-frame and temporal objectives defined in
Section~\ref{sec:training} use the following loss weights: $1$ for $\mathcal{L}_{\theta}$,
$50$ for $\mathcal{L}_{\mathrm{3d}}$, $100$ for $\mathcal{L}_{\mathrm{2d}}$,
$0.1$ for the crop-camera loss, and $10$ each for $\mathcal{L}_{\mathrm{vel}}$
and $\mathcal{L}_{\mathrm{acc}}$. Optimization uses
AdamW~\cite{loshchilov2019adamw} at a fixed learning rate of $2\times10^{-4}$
and weight decay $0.01$. We maintain an exponential moving average of the model
parameters with decay $0.9999$.
Training runs for 210 epochs, corresponding to approximately $200{,}000$
optimization steps. With 16-bit mixed precision on a single GPU, training
takes $13.5\ \mathrm{h}$ on an NVIDIA H200 and $16.5\ \mathrm{h}$ on an NVIDIA GeForce RTX 4090.

\begin{table*}[t]
\scriptsize
\centering
\setlength{\tabcolsep}{4pt}
\caption{
    \textbf{Comparison of expressive whole-body and hand reconstruction} on
    ARCTIC~\cite{fan2023arctic}, MOYO~\cite{tripathi2023moyo}, and
    RICH~\cite{huang2022rich}.
    All errors are per-vertex errors in $\mathrm{mm}$ (lower is better).
    DanceHMR-style fusion denotes our locally implemented and trained baseline,
    not a reproduction of the complete DanceHMR method.
    \modelname\ is best on every hand metric and competitive on full-body metrics.
}
\begin{tabular}{clcccccccccccc}
    \toprule
    & & \multicolumn{4}{c}{ARCTIC} & \multicolumn{4}{c}{MOYO} & \multicolumn{4}{c}{RICH}\\
    \cmidrule(lr){3-6}\cmidrule(lr){7-10}\cmidrule(lr){11-14}
    & & \multicolumn{2}{c}{PA-PVE} & \multicolumn{2}{c}{PVE} & \multicolumn{2}{c}{PA-PVE} & \multicolumn{2}{c}{PVE} & \multicolumn{2}{c}{PA-PVE} & \multicolumn{2}{c}{PVE} \\
    \cmidrule(lr){3-4}\cmidrule(lr){5-6}\cmidrule(lr){7-8}\cmidrule(lr){9-10}\cmidrule(lr){11-12}\cmidrule(lr){13-14}
    & Method &
    All & Hands & All & Hands &
    All & Hands & All & Hands &
    All & Hands & All & Hands \\
    \midrule
    \parbox[t]{4mm}{\multirow{7}{*}{\rotatebox[origin=c]{90}{per-frame}}}
    & AiOS~\cite{sun2024aios} & 30.2 & 19.2 & 47.1 & 38.3 & \(-\) & \(-\) & \(-\) & \(-\) & \(-\) & \(-\) & \(-\) & \(-\) \\
    & Hand4Whole++~\cite{moon2026h4wpp} & \(-\) & \(-\) & 46.0 & 25.0 & \(-\) & \(-\) & \(-\) & \(-\) & \(-\) & \(-\) & \(-\) & \(-\) \\
    & SMPLer-X-H~\cite{cai2023smplerx} & 27.6 & 18.7 & 44.6 & 36.9 & \(-\) & \(-\) & \(-\) & \(-\) & \(-\) & \(-\) & \(-\) & \(-\) \\
    & \quad released checkpoint\textsuperscript{*} & 25.4 & 18.4 & 43.8 & 39.2 & 53.6 & 16.8 & 96.5 & 64.8 & 31.5 & 16.0 & 69.0 & 44.7\\
    & SMPLest-X-H~\cite{yin2025smplest} & 23.2 & 10.7 & 38.1 & 22.6 & \(-\) & \(-\) & \(-\) & \(-\) & \(-\) & \(-\) & \(-\) & \(-\) \\
    & \quad released checkpoint\textsuperscript{*} & 28.1 & 17.8 & 46.1 & 39.0 & 56.8 & 15.7 & 96.0 & 58.3 & 32.0 & 15.9 & 77.1 & 44.5 \\
    & SAM3D-Body\textsuperscript{*}~\cite{yang2026sam3dbody} & 34.7 & 7.1 & 59.5 & 16.3 & 41.6 & 9.9 & 71.4 & 32.9 & 30.2 & 8.2 & 64.7 & 27.5 \\
    \midrule
    \parbox[t]{4mm}{\multirow{3.4}{*}{\rotatebox[origin=c]{90}{temporal}}}
    & DanceHMR (reported)~\cite{shen2026dancehmr} & 28.1 & 8.5 & 44.7 & 24.8 & \(-\) & \(-\) & \(-\) & \(-\) & \(-\) & \(-\) & \(-\) & \(-\) \\
    & DanceHMR-style fusion & \textbf{21.5} & 7.2 & 37.2 & 16.9 & 37.7 & 9.9 & 65.4 & 40.4 & 25.4 & 7.8 & 61.2 & 24.4 \\
    \cmidrule{2-14}
    & \modelname & \textbf{21.5} & \textbf{6.1} & \textbf{35.3} & \textbf{14.7} & \textbf{36.4} & \textbf{8.6} & \textbf{64.2} & \textbf{26.5} & \textbf{24.9} & \textbf{7.7} & \textbf{58.4} & \textbf{21.9} \\
    \bottomrule
\end{tabular}\label{table:expressive_results}\par
\vspace{2pt}
\noindent{\scriptsize
*~Evaluated by us using the publicly released checkpoint.\par}
\end{table*}

\subsubsection{Test-Time Inputs and Preprocessing}
\label{sec:test_time_inputs}
Unless stated otherwise, we obtain \modelname's 23 body-and-foot keypoints with
Sapiens2-1B~\cite{khirodkar2026sapiens2} and treat detections with confidence
below $0.5$ as missing. The detector can be replaced without retraining if it
provides the same 23-keypoint layout.
Camera intrinsics are approximated by setting the focal length to the
full-image diagonal and the
principal point to the image center~\cite{li2022cliff}.
For benchmark sequences with 3D ground truth, we use ground-truth person boxes
and construct each hand box from the projected annotated SMPL or SMPL-X hand
joints. For MOOF, which lacks 3D ground truth, and for arbitrary demonstration
videos, we use detected person boxes. The hand boxes in these cases are
constructed from visible finger keypoints predicted by a whole-body 2D
detector, following HaMeR's demo protocol.
The hand boxes are only used to define separate left- and right-hand crops for the fixed
HaMeR encoder. If no visible finger keypoints are detected for a hand, all
coordinates of the corresponding hand box are set to zero, producing an all-black
input crop for HaMeR. The resulting HaMeR feature is passed to the
initial estimator and HandMR without an explicit hand-visibility indicator.

\subsubsection{Baselines and Comparison Protocol}
\label{sec:baseline_comparison_protocol}
The results for external methods come from the cited publications and our
evaluations of publicly released checkpoints. An asterisk marks the latter.
We evaluate each checkpoint using its official demo or inference code and native
preprocessing with the same benchmark-specific bounding boxes as \modelname.
All local evaluations use the same evaluation code, omit test-time flip
augmentation and temporal smoothing, and use EMA weights for models trained
with EMA.

SAM3D-Body~\cite{yang2026sam3dbody} predicts MHR~\cite{ferguson2025mhr} rather
than SMPL-X parameters. We use the official MHR conversion
code\footnote{\url{https://github.com/facebookresearch/MHR}} to fit SMPL-X
parameters to the MHR output of its locally evaluated
\texttt{sam-\allowbreak{}3d-\allowbreak{}body-\allowbreak{}dinov3} checkpoint.
All locally obtained SAM3D-Body metrics are
computed from the fitted SMPL-X representation.
At the time of evaluation, SAM3D-Body's public release did not include
benchmark evaluation code.

Like \modelname, DanceHMR~\cite{shen2026dancehmr} integrates image features from
hand crops into temporal expressive SMPL-X estimation. Its code and pretrained weights were not
publicly available at evaluation time, so we report its published results
separately and construct a DanceHMR-style fusion baseline. This baseline is
intended as a controlled architectural comparison of body--hand integration,
not a reproduction of the complete method. It shares \modelname's training protocol and Sapiens2-1B
test-time detector and uses the same temporal-transformer backbone as
\modelname's initial estimator. We retain DanceHMR's HaMeR hand features, input
organization, and fusion operation, but replace its HMR2.0 body feature with
\modelname's SAM3D-Body feature.
Separate encoders map the body, left-hand, and right-hand observations
to one per-frame token each. The body observation combines 17 2D body keypoints,
the SAM3D-Body feature, and the same person bounding-box encoding and relative
camera rotation as \modelname. Each hand observation combines 20 2D finger
keypoints, a HaMeR feature, and a separate hand-box encoding. An MLP fuses the
concatenated hand tokens, and its output is added to the body token before the
temporal transformer. \modelname's initial estimator instead integrates both
HaMeR features directly into its per-frame token and uses 23 body-and-foot
keypoints. It receives neither 2D finger
keypoints nor separate hand-box encodings.

\begin{table*}[t]
\scriptsize
\centering
\setlength{\tabcolsep}{4pt}
\caption{
    \textbf{Comparison of hand evidence and integration strategies.}
    Early fusion is an otherwise identical \modelname\ variant with HandMR removed.
    Its deeper and wider variants approximately match \modelname's
    trainable parameter count.
    \modelname\ + HaMeR copy-paste replaces \modelname's final wrist and finger
    rotations with HaMeR predictions.
    Per-vertex errors are in $\mathrm{mm}$ (lower is better).
}
\begin{tabular}{lcccccccccccc}
    \toprule
    & \multicolumn{4}{c}{ARCTIC} & \multicolumn{4}{c}{MOYO} & \multicolumn{4}{c}{RICH}\\
    \cmidrule(lr){2-5}\cmidrule(lr){6-9}\cmidrule(lr){10-13}
    & \multicolumn{2}{c}{PA-PVE} & \multicolumn{2}{c}{PVE} & \multicolumn{2}{c}{PA-PVE} & \multicolumn{2}{c}{PVE} & \multicolumn{2}{c}{PA-PVE} & \multicolumn{2}{c}{PVE} \\
    \cmidrule(lr){2-3}\cmidrule(lr){4-5}\cmidrule(lr){6-7}\cmidrule(lr){8-9}\cmidrule(lr){10-11}\cmidrule(lr){12-13}
    Method &
    All & Hands & All & Hands &
    All & Hands & All & Hands &
    All & Hands & All & Hands \\
    \midrule
    \multicolumn{13}{l}{\emph{Alternative ways of exploiting HaMeR~\cite{pavlakos2024hamer}:}} \\
    \quad early fusion (without HandMR) & 21.8 & 7.2 & 37.4 & 16.4 & 37.5 & 10.3 & \textbf{63.9} & 39.9 & 25.4 & 8.0 & 63.4 & 25.4 \\
    \qquad deeper variant (18 layers) & 21.5 & 7.2 & 35.8 & 16.1 & 37.8 & 10.5 & 64.4 & 40.9 & 25.4 & 8.0 & 63.1 & 25.7 \\
    \qquad wider variant (608 dimensions) & \textbf{21.3} & 7.1 & 36.3 & 16.3 & 37.7 & 10.1 & 64.2 & 40.6 & 25.5 & 8.0 & 63.8 & 25.8 \\
    \quad \modelname\ + HaMeR copy-paste & 22.9 & 8.7 & 37.8 & 26.5 & 38.1 & 9.9 & 65.6 & 40.6 & 26.6 & 8.2 & 59.5 & 37.6 \\
    \midrule
    \multicolumn{13}{l}{\emph{Hand evidence used by both the initial estimator and HandMR:}} \\
    \quad Sapiens2-1B 2D finger keypoints & 22.7 & 8.5 & 38.8 & 18.8 & 37.2 & 10.5 & 64.4 & 34.1 & 25.6 & 8.0 & 61.1 & 24.7 \\
    \quad HaMeR MANO parameters & 22.1 & 6.4 & 37.7 & 15.4 & 37.0 & 9.4 & \textbf{63.9} & 29.5 & \textbf{24.9} & 8.1 & 60.2 & 22.9 \\
    \quad HaMeR image features (\modelname) & 21.5 & \textbf{6.1} & \textbf{35.3} & \textbf{14.7} & \textbf{36.4} & \textbf{8.6} & 64.2 & \textbf{26.5} & \textbf{24.9} & \textbf{7.7} & \textbf{58.4} & \textbf{21.9} \\
    \bottomrule
\end{tabular}\label{table:hand_integration_evidence_ablation}
\end{table*}

\subsection{Expressive Whole-Body and Hand Reconstruction}
\label{sec:expressive_results}

Table~\ref{table:expressive_results} presents our main quantitative comparison
for expressive whole-body reconstruction, with a particular focus on
articulated hands.
\modelname\ obtains the lowest hand PA-PVE and PVE on all three datasets.
Its hand-PVE advantage is largest on MOYO, which contains complex yoga poses
involving extreme wrist articulation.
Compared with the DanceHMR-style baseline, \modelname\ reduces hand PVE by
$13.9\ \mathrm{mm}$ ($40.4$ to $26.5\ \mathrm{mm}$), and compared with SAM3D-Body, the strongest
per-frame competitor, by $6.4\ \mathrm{mm}$ ($32.9$ to $26.5\ \mathrm{mm}$).
The two hand metrics provide complementary views of improvement.
Wrist-aligned hand PVE removes wrist translation but remains strongly affected
by the global wrist orientation.
Per-hand PA-PVE additionally aligns rotation and scale, emphasizing
the local hand configuration.

\modelname\ also obtains the lowest displayed full-body PVE on all three
datasets. For full-body PA-PVE, it ranks first on MOYO and RICH and ties with
the DanceHMR-style baseline on ARCTIC. Finger articulation produces localized
surface displacements, whereas errors in larger body segments such as the
arms and legs move vertices over greater distances. Whole-body averages can
therefore obscure improvements in hand reconstruction. For example, the large MOYO
hand-PVE improvement is accompanied by only a $1.2\ \mathrm{mm}$ reduction in full-body PVE
($65.4$ to $64.2\ \mathrm{mm}$) relative to the DanceHMR-style baseline.

The DanceHMR-style baseline yields lower ARCTIC errors than the published
DanceHMR results. Because the body features, training mixture, and keypoint
detector were changed jointly, and the publication leaves some implementation
and evaluation details unspecified, we cannot isolate the sources of this
discrepancy.

\subsection{HandMR Evaluation and Analysis}

We examine how HandMR affects reconstruction accuracy, temporal stability, and
the initial arm--hand estimate, and how these effects depend on the evidence
representation and refinement design.

\subsubsection{Hand Evidence and Integration Strategies}
\label{sec:hand_refinement_results}

Table~\ref{table:hand_integration_evidence_ablation} examines two design
choices: how hand evidence is integrated and which representation is used.

\paragraph{Dedicated refinement versus early fusion}
To measure the contribution of the HandMR module, we train an otherwise identical
variant of \modelname\ with HandMR removed.
Note that its initial estimator retains the same pair of HaMeR feature vectors.
Since HaMeR features are fused before temporal processing, we denote the variant
as early-fusion baseline.
Adding the HandMR module improves hand PA-PVE and PVE on all three datasets,
with the largest hand-PVE reduction on MOYO ($39.9$ to $26.5\ \mathrm{mm}$).
Across the six full-body columns, \modelname\ is best or
within $0.3\ \mathrm{mm}$ of the best result, indicating that the hand improvements do
not come at a cost to full-body accuracy.
The early-fusion baseline also has similar hand PVE to our DanceHMR-style
baseline (Table~\ref{table:expressive_results}), with differences of at most
$1.0\ \mathrm{mm}$ across the three datasets.

To test whether the hand reconstruction improvements can be explained by
additional model capacity alone, we
increase the early-fusion baseline's trainable parameter count from $45.0$M
to $60.3$M or $58.0$M by deepening or widening its initial estimator,
respectively. These variants approximately match \modelname's $58.9$M
parameters. Although the deeper and wider variants attain lower training
losses than the original early-fusion baseline, neither yields a consistent
improvement on the held-out reconstruction metrics. Both capacity-adjusted
variants have higher hand PA-PVE and PVE than \modelname\ on all three
datasets.

\paragraph{Learned refinement versus direct copy-paste}
The \modelname\ + HaMeR copy-paste variant post-processes the final
\modelname\ prediction by replacing only wrist and finger rotations with
HaMeR outputs, following the integration strategy of
FrankMocap~\cite{rong2021frankmocap}. It uses the same trained \modelname\
checkpoint, fixed HaMeR checkpoint, hand crops, and inference pass. The
handedness correction and rotation conversion are detailed in the
supplementary material (Section~\ref{sec:hamer_copy_paste_protocol}).
This copy-paste variant produces substantially higher errors than \modelname\ on
both hand metrics for all three datasets. On MOYO, hand PVE increases from
$26.5$ to $40.6\ \mathrm{mm}$. HaMeR predicts each hand from its crop independently.
The temporal context and initial arm--hand state available to HandMR may
help resolve ambiguities under partial occlusion.

\begin{figure*}[t]
    \centering
    \includegraphics[width=\textwidth]{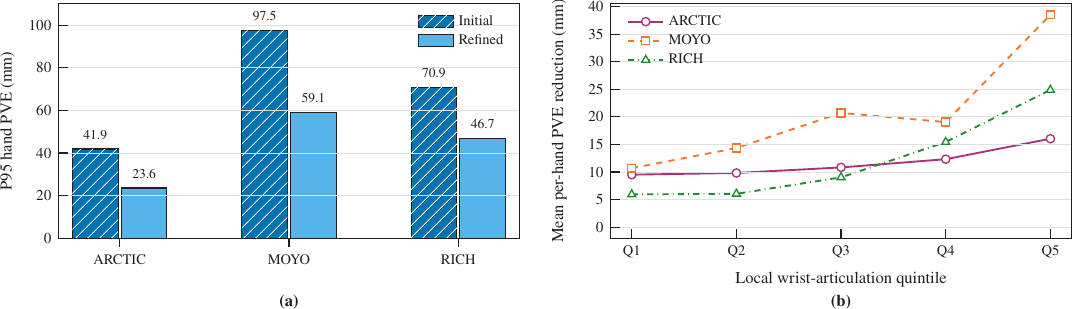}
    \caption{
        \textbf{Where HandMR refines the initial estimate.}
        Both panels compare initial and HandMR-refined outputs from one
        \modelname\ checkpoint on the same evaluated frames.
        \textbf{(a) Severity of high-error predictions.}
        The 95th percentile (P95) of wrist-aligned hand PVE after averaging
        the two hands' errors per frame. Percentiles are computed separately
        for initial and refined outputs.
        \textbf{(b) Error reduction versus wrist articulation.}
        Mean per-hand PVE reduction (initial minus refined error) by
        within-dataset quintile of ground-truth local wrist articulation
        (Q1: lowest, Q5: highest). Articulation is the $\mathrm{SO}(3)$ geodesic
        distance from the elbow-relative wrist rotation to identity.
        Each hand's articulation is paired with its PVE reduction.
        Positive values indicate improvement.
    }\label{fig:refinement_distribution}
\end{figure*}

\paragraph{Hand-evidence representation}
The lower block compares \modelname\ with two variants that replace HaMeR
image features in both the initial estimator and HandMR. The
2D-finger-keypoint variant uses 20 2D finger keypoints per hand, and the
MANO-parameter variant uses HaMeR's explicit wrist-orientation and finger-pose
estimates. During training, we synthesize the 2D finger keypoints from
ground-truth SMPL-X joints using the procedure in Section~\ref{sec:training}.
At test time, Sapiens2-1B~\cite{khirodkar2026sapiens2} detects them. Because the synthetic
training coordinates can be regenerated under an augmented orientation, we
also apply FootMR's random global-orientation augmentation when constructing
the keypoint variant's HandMR inputs. For both variants, we jointly
retrain the initial estimator, FootMR, and HandMR from scratch.

MANO parameters reduce hand PVE relative to 2D keypoints on every dataset.
Within our pipeline, HaMeR image features perform better still, obtaining the
lowest hand PA-PVE and PVE on all three datasets. For example, MOYO hand PVE
is $34.1\ \mathrm{mm}$ with 2D keypoints, $29.5\ \mathrm{mm}$ with MANO parameters, and $26.5\ \mathrm{mm}$
with image features. The latent features may retain appearance cues related
to visibility and pose ambiguity that the temporal models can combine with
motion and body context.
Adding MANO parameters or 2D finger keypoints to the HaMeR image
features did not further improve reconstruction accuracy in our experiments.

\subsubsection{Temporal Accuracy and Stability}
\label{sec:temporal_hand_reconstruction}

\begin{table}[t]
\scriptsize
\centering
\caption{
    \textbf{Temporal accuracy and stability of hand reconstruction.}
    Metrics use 21 wrist-centered joints per hand.
    MPJVE ($\mathrm{mm/s}$) and Accel ($\mathrm{m/s^2}$) are
    ground-truth-referenced; Jitter ($10\ \mathrm{m/s^3}$) is prediction-only.
    \modelname\ + HaMeR copy-paste replaces \modelname's wrist and finger
    rotations with independent per-frame HaMeR predictions.
    The final two rows are temporal methods.
    DanceHMR-style fusion denotes our locally implemented and trained baseline.
}
\setlength{\tabcolsep}{3pt}
\begin{tabular}{lcccccc}
    \toprule
    & \multicolumn{3}{c}{ARCTIC} & \multicolumn{3}{c}{MOYO}\\
    \cmidrule(lr){2-4}\cmidrule(lr){5-7}
    Method & MPJVE & Accel & Jitter & MPJVE & Accel & Jitter \\
    \midrule
    SMPLer-X-H\textsuperscript{*}~\cite{cai2023smplerx} & 204.5 & 8.4 & 43.2 & 196.9 & 8.3 & 45.1 \\
    SMPLest-X-H\textsuperscript{*}~\cite{yin2025smplest} & 206.5 & 8.5 & 43.6 & 187.4 & 7.8 & 42.2 \\
    SAM3D-Body\textsuperscript{*}~\cite{yang2026sam3dbody} & 129.4 & 6.0 & 32.4 & 208.6 & 10.5 & 53.4 \\
    \modelname\ + HaMeR copy-paste~\cite{pavlakos2024hamer} & 179.2 & 7.7 & 40.9 & 260.3 & 12.4 & 68.1 \\
    \midrule
    DanceHMR-style fusion~\cite{shen2026dancehmr} & 65.2 & 1.6 & \textbf{4.6} & 58.8 & \textbf{1.2} & \textbf{3.5} \\
    \modelname& \textbf{57.9} & \textbf{1.5} & 5.2 & \textbf{51.3} & 1.4 & 5.2 \\
    \bottomrule
\end{tabular}\label{table:temporal_stability}\par
\vspace{2pt}
\noindent{\scriptsize
*~Evaluated by us using the publicly released checkpoint.\par}
\end{table}

We next assess whether the spatial improvements are accompanied by accurate
and stable hand dynamics. Table~\ref{table:temporal_stability} evaluates
wrist-centered hand trajectories on ARCTIC and MOYO. MPJVE and acceleration
error measure fidelity to the annotated dynamics, whereas prediction-only
jitter measures smoothness. These temporal metrics do not establish spatial
reconstruction quality in isolation and should be interpreted alongside
framewise hand errors.

On both datasets, \modelname\ and the DanceHMR-style baseline obtain
substantially lower MPJVE, acceleration error, and jitter than the per-frame
methods and the \modelname\ + HaMeR copy-paste baseline.
\modelname\ achieves the lowest MPJVE among the evaluated methods.
Compared with the DanceHMR-style baseline, it obtains $57.9$ versus
$65.2\ \mathrm{mm/s}$ on ARCTIC and $51.3$ versus $58.8\ \mathrm{mm/s}$ on MOYO.
The DanceHMR-style baseline has lower
jitter on both datasets. Acceleration error favors \modelname\ on ARCTIC
and the baseline on MOYO.

\begin{figure*}[t]
    \centering
    \includegraphics{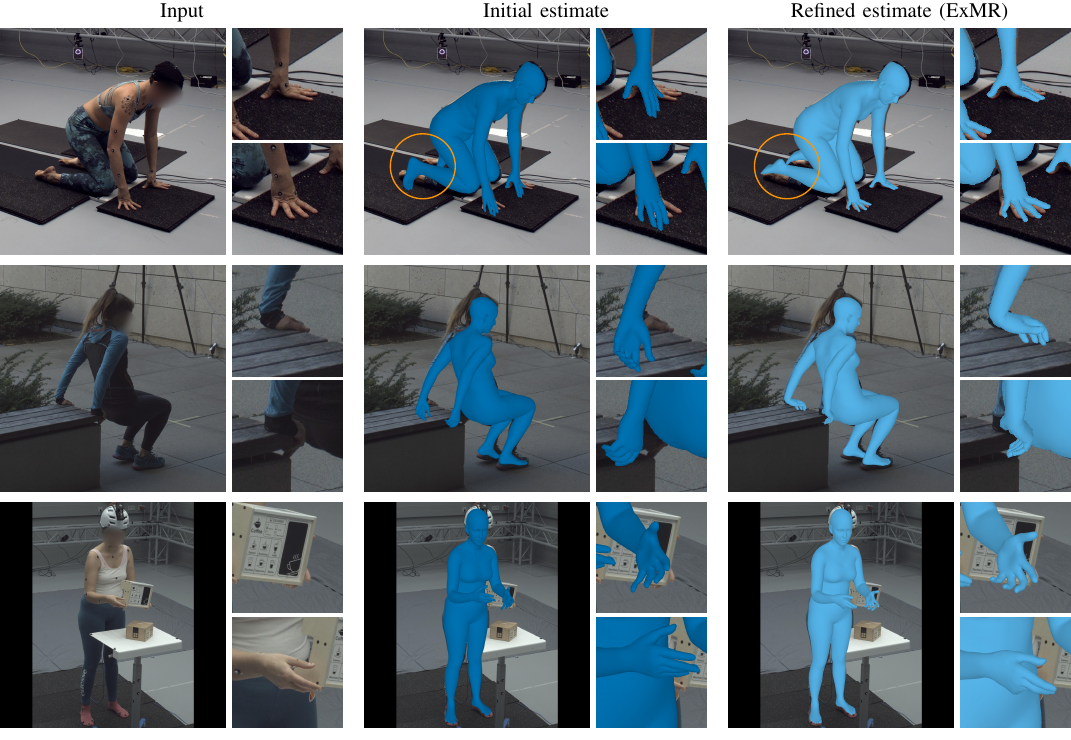}\par\prevdepth=175.55128pt\relax
    \caption{
        \textbf{Qualitative examples of extremity refinement in \modelname.}
        Rows show one example each from MOYO (top), RICH (middle), and ARCTIC
        (bottom). Initial and refined estimates are outputs
        of the same \modelname\ checkpoint.
        Amber circles in the MOYO row highlight the correction
        of the initial foot orientation.
    }\label{fig:qualitative_results}
\end{figure*}

\subsubsection{Refinement Behavior}
\label{sec:where_handmr_refines}

We analyze two aspects of HandMR's behavior that mean reconstruction errors
alone do not describe: whether refinement reduces the severity of high-error
predictions and how its gains vary with wrist articulation. Both analyses
compare \modelname's initial estimates with its final HandMR-refined outputs
from the same checkpoint on the same evaluated frames.

To assess high-error predictions, we compare the 95th percentile (P95) of
the per-frame mean hand PVE before and after refinement. This summarizes
the upper tail of the error distribution.
Fig.~\ref{fig:refinement_distribution}(a) shows that P95 decreases on all
three datasets: from $41.9$ to $23.6\ \mathrm{mm}$ on ARCTIC, from $97.5$ to $59.1\ \mathrm{mm}$
on MOYO, and from $70.9$ to $46.7\ \mathrm{mm}$ on RICH.

We also examine whether more articulated wrists receive larger error
reductions. MOYO motivates this analysis: its yoga poses involve highly
articulated wrists, and HandMR's hand-PVE gain over early fusion is largest
on this dataset. To examine
the refinement stage, we group paired per-hand PVE reductions between the
initial and refined outputs by ground-truth local
wrist articulation (Fig.~\ref{fig:refinement_distribution}(b)).
The highest-articulation quintile has the largest mean reduction on every
dataset. MOYO shows the largest difference between the lowest and highest
quintiles, with mean reductions of $10.7$ and $38.5\ \mathrm{mm}$, respectively.

Fig.~\ref{fig:qualitative_results} shows qualitative results which
highlight the changes from \modelname's initial estimates to its refined outputs.
Additional qualitative comparisons with SAM3D-Body are provided in
Fig.~\ref{fig:sam3d_comparison} of the supplementary material.
The accompanying supplementary video illustrates reconstructed full-body
motion over time.

\subsubsection{HandMR Design Choices}
\label{sec:handmr_design_analysis}

\begin{table}[!t]
\scriptsize
\centering
\caption{\textbf{Design ablations of the HandMR module.}
    All configurations are retrained from scratch.
    Top: joint-rotation inputs. The All column adds all remaining body joints.
    Finger rotations are parent-relative and all others global.
    Fingers and wrists are predicted residually given initial rotations
    and directly otherwise. Elbows are refined residually only
    when supplied.
    Bottom: variants of the final \modelname\ configuration
    (\(\blacktriangleright\)).
    Hand PA-PVE and PVE are in $\mathrm{mm}$ (lower is better).
}\label{table:ablation_handmr_design}
    \setlength{\tabcolsep}{4pt}
    \begin{tabular}{@{}ccccccccc@{}}
        \toprule
        & \multicolumn{4}{c}{Input Joint Rotations} & \multicolumn{2}{c}{ARCTIC} & \multicolumn{2}{c}{MOYO} \\
        \cmidrule(lr){2-5}\cmidrule(lr){6-7}\cmidrule(lr){8-9}
        & Finger & Wrist & Elbow & All & PA-PVE & PVE & PA-PVE & PVE \\
        \midrule
        & \(\times\) & \(\times\) & \(\times\) & \(\times\) & 6.6 & 16.6 & 9.0 & 31.5 \\
        & \(\surd\) & \(\times\) & \(\times\) & \(\times\) & 6.3 & 14.8 & 8.8 & 27.5 \\
        & \(\times\) & \(\surd\) & \(\times\) & \(\times\) & 7.4 & 15.1 & 9.8 & 28.5 \\
        & \(\surd\) & \(\surd\) & \(\times\) & \(\times\) & 6.0 & 14.7 & 8.6 & 26.7 \\
        \(\blacktriangleright\) & \(\surd\) & \(\surd\) & \(\surd\) & \(\times\) & 6.1 & 14.7 & 8.6 & 26.5 \\
        & \(\surd\) & \(\surd\) & \(\surd\) & \(\surd\) & 6.2 & 14.6 & 8.6 & 26.5\\
        \midrule
        \multicolumn{5}{@{}l}{\(\blacktriangleright\) + SAM3D-Body feature input} & 6.3 & 14.7 & 9.1 & 27.8 \\
        \multicolumn{5}{@{}l}{\(\blacktriangleright\) with elbow-relative wrist rotations} & 6.1 & 14.6 & 8.9 & 30.4 \\
        \multicolumn{5}{@{}l}{\(\blacktriangleright\) without elbow refinement} & 6.1 & 14.6 & 8.5 & 26.6 \\
        \bottomrule
    \end{tabular}
\end{table}

We finally analyze three aspects of HandMR's design: its inputs,
wrist-rotation representation, and elbow refinement.

\begin{table*}[t]
    \scriptsize
    \centering
    \caption{
        \textbf{Effect of elbow refinement on elbow and wrist rotation accuracy.}
        Both rows jointly retrain the initial estimator and the FootMR and
        HandMR modules from scratch. They differ only in whether HandMR
        predicts residual updates to the elbows. G-RE and L-RE denote global camera-frame and parent-relative
        geodesic rotation errors, respectively, in degrees.
    }\label{table:elbow_refinement}
    \setlength{\tabcolsep}{4pt}
    \begin{tabular}{lcccccccccccc}
        \toprule
        & \multicolumn{4}{c}{ARCTIC} & \multicolumn{4}{c}{MOYO} & \multicolumn{4}{c}{RICH} \\
        \cmidrule(lr){2-5}\cmidrule(lr){6-9}\cmidrule(lr){10-13}
        & \multicolumn{2}{c}{Elbow} & \multicolumn{2}{c}{Wrist}
        & \multicolumn{2}{c}{Elbow} & \multicolumn{2}{c}{Wrist}
        & \multicolumn{2}{c}{Elbow} & \multicolumn{2}{c}{Wrist} \\
        \cmidrule(lr){2-3}\cmidrule(lr){4-5}
        \cmidrule(lr){6-7}\cmidrule(lr){8-9}
        \cmidrule(lr){10-11}\cmidrule(lr){12-13}
        Method & G-RE & L-RE & G-RE & L-RE
        & G-RE & L-RE & G-RE & L-RE
        & G-RE & L-RE & G-RE & L-RE \\
        \midrule
        without elbow refinement
        & 8.3 & 9.0 & \textbf{10.9} & 11.4
        & 18.4 & 17.5 & \textbf{19.4} & 21.2
        & 18.4 & 16.3 & 17.3 & 20.0 \\
        with elbow refinement (\modelname)
        & \textbf{7.4} & \textbf{8.4} & 11.0 & \textbf{10.4}
        & \textbf{16.9} & \textbf{15.9} & \textbf{19.4} & \textbf{19.6}
        & \textbf{16.5} & \textbf{15.4} & \textbf{17.1} & \textbf{18.4} \\
        \bottomrule
    \end{tabular}
\end{table*}

\begin{table*}[t]
\scriptsize
\centering
\caption{
    \textbf{Comparison of foot-pose reconstruction} on
    MOYO~\cite{tripathi2023moyo}, RICH~\cite{huang2022rich}, and
    MOOF~\cite{wehrbein26footmr}. FootMR (conference) and FootMR
    (revised) denote complete body-and-foot systems. Both use the same FootMR
    ankle-refinement module design. Their corresponding
    GVHMR\textsubscript{23j} baselines use matched training and inference
    configurations without the FootMR module. \modelname\ adds the HandMR module to the revised system. AJAE is in degrees,
    N-MPJPE\textsubscript{F} in $\mathrm{mm}$, PCK\textsubscript{F} in $\%$,
    and N-FKE\textsubscript{2d} in normalized pixel space.
}
\begin{tabular}{clcccccc}
    \toprule
    & & \multicolumn{2}{c}{MOYO} & \multicolumn{2}{c}{RICH} & \multicolumn{2}{c}{MOOF}\\
    \cmidrule(lr){3-4}\cmidrule(lr){5-6}\cmidrule(lr){7-8}
    & Method &
    AJAE\(\downarrow\) & N-MPJPE\textsubscript{F}\(\downarrow\) & AJAE\(\downarrow\) & N-MPJPE\textsubscript{F}\(\downarrow\) & PCK\textsubscript{F} 0.05\(\uparrow\) & N-FKE\textsubscript{2d}\(\downarrow\) \\
    \midrule
    \parbox[t]{4mm}{\multirow{5}{*}{\rotatebox[origin=c]{90}{per-frame}}}
    & HMR2.0\textsuperscript{*}~\cite{goel2023humans} & 42.0 & 55.7 & \(-\) & 36.9 & 76.1 & 1.95 \\
    & ReFit\textsuperscript{*}~\cite{wang2023refit} & 41.8 & 47.6 & \(-\) & 30.3 & 81.8 & 1.68 \\
    & TokenHMR\textsuperscript{*}~\cite{dwivedi2024tokenhmr} & 38.0 & 48.1 & \(-\) & 27.7 & 74.5 & 1.85 \\
    & CameraHMR\textsuperscript{*}~\cite{patel2024camerahmr} & 31.0 & 39.1 & \(-\) & 24.8 & 88.3 & 1.42 \\
    & SAM3D-Body\textsuperscript{*}~\cite{yang2026sam3dbody} & 24.4 & 32.6 & 17.5 & 21.9 & 96.6 & 0.89 \\
    \midrule
    \parbox[t]{4mm}{\multirow{9}{*}{\rotatebox[origin=c]{90}{temporal}}}
    & WHAM\textsuperscript{*}~\cite{shin2024wham} & 41.7 & 49.1 & \(-\) & 28.3 & 68.1 & 1.94 \\
    & TRAM\textsuperscript{*}~\cite{wang2024tram} & 39.6 & 47.3 & \(-\) & 26.5 & 68.0 & 1.60 \\
    & GVHMR\textsuperscript{*}~\cite{shen2024gvhmr} & 37.3 & 45.9 & 18.0 & 27.3 & 76.1 & 1.70 \\
    \cmidrule{2-8}
    & GVHMR\textsubscript{23j} (conference baseline)~\cite{wehrbein26footmr} & 29.6 & 37.5 & 16.2 & 24.2 & 89.2 & 1.06 \\
    & FootMR (conference)~\cite{wehrbein26footmr} & 25.9 & 32.4 & 15.8 & 23.0 & 92.6 & \textbf{0.67} \\
    \cmidrule{2-8}
    & GVHMR\textsubscript{23j} (revised baseline) & 25.3 & 32.2 & 13.3 & 19.7 & 95.9 & 0.84 \\
    & FootMR (revised) & 20.4 & 25.7 & \textbf{12.6} & \textbf{18.7} & \textbf{97.5} & \textbf{0.67}  \\
    & \modelname & \textbf{20.0} & \textbf{25.4} & 12.8 & \textbf{18.7} & 97.3 & \textbf{0.67} \\
    \bottomrule
\end{tabular}\label{table:foot_reconstruction_results}\par
\vspace{2pt}
\noindent{\scriptsize *~Evaluated by us using the publicly released checkpoint.\par}
\end{table*}

\paragraph{Inputs and prediction formulation}
Table~\ref{table:ablation_handmr_design} varies the initial joint
rotations supplied to HandMR.
For fingers and wrists, supplying an initial rotation also changes prediction
from direct to residual.
Initial finger rotations reduce hand PVE more than wrist rotations alone on
both datasets. Supplying both improves hand PA-PVE and PVE, reducing MOYO hand PVE from $31.5$ to
$26.7\ \mathrm{mm}$.
The hand errors differ by at most $0.2\ \mathrm{mm}$ from those of the final \modelname\
configuration, which also supplies and refines the elbows.
Initial wrist and finger estimates thus complement HaMeR features for hand reconstruction.
We hypothesize that predictions from the whole-body temporal
initial estimator provide useful context when local hand evidence is
ambiguous, for example under occlusion. Residual prediction can preserve
these estimates when no correction is required.

Supplying all remaining initial body-joint rotations yields no consistent
improvement in hand accuracy. Adding the SAM3D-Body feature, already used by
the initial estimator, gives similar performance on ARCTIC but increases
MOYO hand PA-PVE and PVE by $0.5$ and $1.3\ \mathrm{mm}$, respectively. These results
support retaining HandMR's compact input set.

\paragraph{Wrist-rotation representation}
Using elbow-relative rather than global wrist rotations leaves both ARCTIC hand
metrics essentially unchanged but increases MOYO hand PVE from $26.5$ to
$30.4\ \mathrm{mm}$.
Following the analogous finding for FootMR~\cite{wehrbein26footmr}, we
hypothesize that global wrist rotations span a broader training distribution
than elbow-relative rotations, aiding generalization to MOYO's extreme wrist
articulation.

\paragraph{Effect of elbow refinement}
Disabling elbow refinement changes hand PA-PVE and PVE by at most $0.1\ \mathrm{mm}$
on ARCTIC and MOYO (Table~\ref{table:ablation_handmr_design}), so we evaluate
elbow and wrist rotation accuracy directly in
Table~\ref{table:elbow_refinement}.
Across all three datasets, elbow refinement reduces elbow G-RE by
approximately $8\%$--$11\%$, elbow L-RE by $6\%$--$9\%$, and wrist L-RE by
$8\%$--$9\%$ (absolute reductions of $0.6$--$1.9^\circ$).
Wrist G-RE changes by at most $0.2^\circ$, with no consistent improvement.
As described in Section~\ref{sec:hand_refinement}, SMPL-X's parent-relative
wrist rotation is recovered from the global wrist and elbow estimates.
Refining the elbow estimate therefore improves elbow--wrist consistency,
even though global wrist accuracy and hand-surface errors remain comparable.

\begin{table}[!b]
\scriptsize
\centering
\caption{
    \textbf{Cumulative changes from FootMR (conference) to FootMR (revised).}
    The FootMR module retains its design and is jointly retrained
    with the initial estimator for each training configuration.
    The detector change affects only test-time inputs.
    Final row: FootMR (revised) in Table~\ref{table:foot_reconstruction_results}.
    EMA: exponential moving average of model parameters.
    AJAE in degrees and PCK\textsubscript{F}~in~$\%$.
}\label{table:footmr_revision_analysis}
    \begin{tabular}{@{}lccc@{}}
        \toprule
        & MOYO & RICH & MOOF \\
        \cmidrule(lr){2-4}
        Configuration & AJAE\(\downarrow\) & AJAE\(\downarrow\) & PCK\textsubscript{F} 0.05\(\uparrow\) \\
        \midrule
        FootMR (conference)~\cite{wehrbein26footmr} & 25.9 & 15.8 & 92.6 \\
        \(+\) SAM3D-Body feature \& training settings & 22.5 & 14.3 & 95.0 \\
        \(+\) revised training data & 22.1 & 13.7 & 96.2 \\
        \(+\) revised losses \& reweighting & 21.4 & 13.7 & 97.3 \\
        \(+\) EMA & 21.0 & 13.1 & 97.5 \\
        \(+\) Sapiens2-1B keypoints (test time) & 20.4 & 12.6 & 97.5 \\
        \bottomrule
    \end{tabular}
\end{table}

\begin{table*}[t]
\scriptsize
\centering
\caption{
    \textbf{Comparison of body-only reconstruction} on 3DPW~\cite{marcard2018eccv}, EMDB-1~\cite{kaufmann2023emdb}, and RICH~\cite{huang2022rich}.
    Evaluation uses the default hand pose for \modelname\ and the locally evaluated
    SAM3D-Body checkpoint.
    PA-MPJPE, MPJPE, and PVE are in $\mathrm{mm}$, and Accel is in
    $\mathrm{m/s^2}$.
    Parentheses give the number of joints used for PA-MPJPE, MPJPE, and Accel.
}
\begin{tabular}{clcccccccccccc}
    \toprule
    & & \multicolumn{4}{c}{3DPW (14)} & \multicolumn{4}{c}{EMDB-1 (24)} & \multicolumn{4}{c}{RICH (24)}\\
    \cmidrule(lr){3-6}\cmidrule(lr){7-10}\cmidrule(lr){11-14}
    & Method &
    PA-MPJPE & MPJPE & PVE & Accel &
    PA-MPJPE & MPJPE & PVE & Accel &
    PA-MPJPE & MPJPE & PVE & Accel \\
    \midrule
    \parbox[t]{4mm}{\multirow{7}{*}{\rotatebox[origin=c]{90}{per-frame}}}
    & HMR2.0~\cite{goel2023humans} & 44.4 & 69.8 & 82.2 & 18.1 & 60.6 & 98.0 & 120.3 & 19.8 & 48.1 & 96.0 & 110.9 & 18.8 \\
    & ReFit~\cite{wang2023refit} & 40.5 & 65.3 & 75.1 & 18.5 & 58.6 & 88.0 & 104.5 & 20.7 & 47.9 & 80.7 & 92.9 & 17.1 \\
    & TokenHMR~\cite{dwivedi2024tokenhmr} & 44.3 & 71.0 & 84.6 & \(-\) & 55.6 & 91.7 & 109.4 & \(-\) & \(-\) & \(-\) & \(-\) & \(-\) \\
    & CameraHMR~\cite{patel2024camerahmr} & 35.1 & 56.0 & 65.9 & \(-\) & 43.3 & 70.2 & 81.7 & \(-\) & 34.0 & 55.7 & 64.4 & \(-\) \\
    & NLF-L +fit~\cite{sarandi2024nlf} & 33.6 & 54.9 & 63.7 & \(-\) & 40.9 & 68.4 & 80.6 & \(-\) & \(-\) & \(-\) & \(-\) & \(-\) \\
    & SAM3D-Body~\cite{yang2026sam3dbody} & 33.8 & 54.8 & 63.6 & \(-\) & 38.2 & \textbf{61.7} & \textbf{72.5} & \(-\) & 30.9 & 53.7 & 60.3 & \(-\) \\
    & \quad released checkpoint\textsuperscript{*} & 35.1 & 55.5 & 65.4 & 14.0 & 39.6 & 64.5 & 76.9 & 15.1 & 33.0 & 56.4 & 62.8 & 10.6 \\
    \midrule
    \parbox[t]{4mm}{\multirow{8}{*}{\rotatebox[origin=c]{90}{temporal}}}
    & WHAM~\cite{shin2024wham} & 35.9 & 57.8 & 68.7 & 6.6 & 50.4 & 79.7 & 94.4 & 5.3 & 44.3 & 80.0 & 91.2 & 5.3 \\
    & TRAM~\cite{wang2024tram} & 35.6 & 59.3 & 69.6 & 4.9 & 45.7 & 74.4 & 86.6 & 4.9 & \(-\) & \(-\) & \(-\) & \(-\) \\
    & GVHMR~\cite{shen2024gvhmr} & 36.2 & 55.6 & 67.2 & 5.0 & 42.7 & 72.6 & 84.2 & 3.6 & 39.5 & 66.0 & 74.4 & 4.1 \\
    & PromptHMR-Vid~\cite{wang2025prompthmr} & 35.5 & 56.9 & 67.3 & \(-\) & 40.1 & 68.1 & 79.2 & \(-\) & 37.0 & 57.4 & 65.8 & \(-\) \\
    & GENMO~\cite{li2025genmo} & 34.6 & 53.9 & 65.8 & 5.2 & 42.5 & 73.0 & 84.8 & 3.8 & 39.1 & 66.8 & 75.4 & 4.1 \\
    & DuoMo~\cite{wang2026duomo} & \(-\) & \(-\) & \(-\) & \(-\) & 41.8 & 67.1 & 78.2 & \(-\) & 34.8 & 51.4 & 58.9 & \(-\) \\
    & DanceHMR~\cite{shen2026dancehmr} & 35.2 & 53.6 & 65.1 & 4.9 & 40.1 & 64.3 & 75.4 & 3.5 & \(-\) & \(-\) & \(-\) & \(-\) \\
    \cmidrule{2-14}
    & \modelname& \textbf{33.0} & \textbf{51.8} & \textbf{60.4} & \textbf{4.1} & \textbf{37.4} & 64.1 & 74.6 & \textbf{3.3} & \textbf{27.3} & \textbf{50.0} & \textbf{55.7} & \textbf{3.6} \\
    \bottomrule
\end{tabular}\label{table:body_only_results}\par
\vspace{2pt}
\noindent{\scriptsize *~Evaluated by us using the publicly released checkpoint.\par}
\end{table*}

\subsection{Foot Reconstruction}
\label{sec:foot_results}

\subsubsection{Effect of Dedicated Foot Refinement Under the Revised Configuration}

We test whether dedicated foot refinement remains beneficial when the
initial estimator itself becomes more accurate. Our revised initial
estimator uses image features from SAM3D-Body~\cite{yang2026sam3dbody}, which
explicitly targets accurate foot reconstruction. It also uses revised
training data including BEDLAM2~\cite{tesch2025bedlam2}.
These changes may improve the initial foot estimates enough to make the
FootMR module introduced in our conference paper~\cite{wehrbein26footmr}
redundant.

We denote the published body-and-foot system as \emph{FootMR (conference)}
and its revised configuration as \emph{FootMR (revised)}.
The corresponding baselines,
GVHMR\textsubscript{23j} (conference baseline) and
GVHMR\textsubscript{23j} (revised baseline), match the training and inference
configurations of their respective systems but omit the FootMR module.
FootMR (revised) is a configuration of \modelname\ without the hand-specific
components: HaMeR features, finger outputs and losses, and HandMR.
We retrain FootMR jointly with the initial estimator while retaining the
module's design from the conference paper. All remaining training and
inference settings match those of \modelname.

Table~\ref{table:foot_reconstruction_results} reports 3D foot accuracy on MOYO
and RICH and 2D reprojection accuracy on MOOF. The revised
GVHMR\textsubscript{23j} baseline improves all six metrics over the conference
baseline. Adding the FootMR module under the revised configuration further
improves all six metrics. On MOYO, FootMR (revised) reduces AJAE from
$25.3^\circ$ to $20.4^\circ$ ($19\%$) and
N-MPJPE\textsubscript{F} from $32.2$ to $25.7\ \mathrm{mm}$ ($20\%$). On RICH, the
reductions are approximately $5\%$ for both metrics. On MOOF, it improves
PCK\textsubscript{F} from $95.9\%$ to $97.5\%$ and reduces
N-FKE\textsubscript{2d} from $0.84$ to $0.67$.
These results show that the FootMR module remains beneficial under the
revised configuration, even with more accurate initial foot estimates.
The largest 3D gains from
the FootMR module occur on MOYO, mirroring the pattern observed when
adding the HandMR module in the preceding subsections.

For each of the six metrics, FootMR (revised) and \modelname\ also
outperform the best external result reported in the table.
Compared with SAM3D-Body, the strongest per-frame method evaluated here,
FootMR (revised) reduces MOYO AJAE from $24.4^\circ$ to $20.4^\circ$ and
N-MPJPE\textsubscript{F} from $32.6$ to $25.7\ \mathrm{mm}$.
Finally, \modelname\ and FootMR (revised) differ by at most
$0.4^\circ$ in AJAE, $0.3\ \mathrm{mm}$ in N-MPJPE\textsubscript{F}, and
$0.2$ percentage points in PCK\textsubscript{F}.
The small differences indicate that adding articulated-hand estimation and
the HandMR module preserves foot reconstruction accuracy, consistent with
FootMR and HandMR using different evidence and updating disjoint joint
groups. FootMR's contribution within the complete \modelname\ system is also
visible in the MOYO example in the top row of
Fig.~\ref{fig:qualitative_results}, where the module substantially corrects
the initial foot orientation.

\subsubsection{From FootMR (conference) to FootMR (revised)}
Compared with FootMR (conference), FootMR (revised) improves both 3D metrics on
MOYO and RICH and PCK\textsubscript{F} on MOOF
(Table~\ref{table:foot_reconstruction_results}).
Table~\ref{table:footmr_revision_analysis} traces this difference through
cumulative system-level revisions made while holding the FootMR design fixed.
Each configuration builds on the preceding one, so the comparison does not
isolate independent effects or interactions.
We report AJAE on MOYO and RICH and PCK\textsubscript{F} on MOOF.
Table~\ref{table:foot_reconstruction_results} reports all metrics for the two
endpoints.

The first stage replaces the HMR2.0 image feature with the SAM3D-Body feature
and adopts \modelname's training settings: a batch size of 128 instead of 256,
a fixed learning rate, and 210 rather than 500 epochs.
We observed no meaningful accuracy change from these practical adjustments
and therefore do not treat them as separate ablation factors.
The data stage adds BEDLAM2 and ARCTIC and removes AMASS relative to the
conference training mixture. The loss stage replaces the geometric
$\ell_2$ losses with $\ell_1$ losses, adds velocity and acceleration losses,
and adjusts the loss weights.
The next stage introduces an exponential moving average (EMA) of the model
parameters.
At test time, FootMR (conference) obtains the 17 body keypoints with
ViTPose~\cite{xu22vitpose} and the six foot keypoints with
Sapiens-2B~\cite{khirodkar2024sapiens}.
The final stage replaces both detectors with
Sapiens2-1B~\cite{khirodkar2026sapiens2} to obtain all 23 keypoints.
Training continues to use keypoints synthesized from ground-truth motion.

Across the cumulative sequence, MOYO AJAE decreases from $25.9^\circ$ to
$20.4^\circ$, RICH AJAE from $15.8^\circ$ to $12.6^\circ$, and MOOF
PCK\textsubscript{F} increases from $92.6\%$ to $97.5\%$.
The supplementary material (Section~\ref{sec:footmr_design_analysis}) evaluates
FootMR's inherited joint-context, rotation-representation, and augmentation
design choices within the complete \modelname\ system.

\subsection{Body-Only Reconstruction}
\label{sec:body_only_reconstruction}

We complement the hand and foot evaluations by assessing
\modelname\ under standard body-reconstruction protocols on 3DPW,
EMDB-1, and RICH (Table~\ref{table:body_only_results}).
As detailed in Section~\ref{sec:experimental_protocol}, we use default
finger-joint rotations in both predictions and ground truth when constructing
the evaluation meshes. FootMR's ankle refinements and HandMR's elbow and
wrist refinements remain active.

Among the reported results, \modelname\ has the lowest PA-MPJPE, MPJPE, and
PVE on 3DPW and RICH, and the lowest PA-MPJPE on EMDB-1.
For example, its RICH PVE is $55.7\ \mathrm{mm}$, compared with $58.9\ \mathrm{mm}$ for DuoMo,
the next-lowest reported value.
On EMDB-1, \modelname's MPJPE and PVE are the lowest among temporal methods,
while the published SAM3D-Body numbers are lower.
For temporal accuracy, \modelname\ achieves the lowest reported acceleration
error on all three datasets.
The methods differ in their training data, image features, objectives,
inference settings, and potentially their evaluation details, so this
comparison does not isolate the source of the body-only differences.
Taken together with the preceding hand- and foot-specific evaluations,
these results show that \modelname\ combines
competitive body-only benchmark accuracy with fine-grained hand and foot
reconstruction.

\section{Conclusion}
\modelname\ combines whole-body motion estimation with dedicated temporal refinement
of the hands and feet. It extends our conference body-and-foot system with HandMR,
which combines specialist hand image features with initial arm--hand rotations to
refine elbows, wrists, and fingers. Paired with FootMR's
keypoint-based ankle refinement, this yields a jointly trained system
for reconstructing body, hand, and foot motion from monocular video.
FootMR and HandMR follow a common principle: revisiting local
observations together with explicit joint estimates to predict residual rotation
corrections over time.
Across the evaluated benchmarks, \modelname\ outperforms all competitors on
hand and foot reconstruction while achieving competitive body accuracy.
The experiments also show that how local evidence is used matters beyond its availability.
In matched comparisons, adding the HandMR and FootMR modules improves hand and foot
reconstruction even though the initial estimator already receives the same evidence.
These findings suggest that dedicated temporal refinement improves extremity reconstruction
by exploiting local evidence underused by a shared whole-body representation.

\modelname\ processes each person independently, without explicitly modeling
interactions with other people or objects. It does not
enforce collision-free geometry or physically consistent grasps and object
contacts, and reconstructed meshes may contain self-intersections.
Facial pose and expression also remain outside the scope of this work.
Future work could incorporate self-collision constraints and jointly estimate
human motion, object geometry, and contact. Our quantitative evaluation
focuses on camera-frame reconstruction. Investigating whether improved
extremity estimates benefit world-space motion recovery and foot--ground
contact consistency is another promising direction.

\bibliographystyle{exmrIEEE}
\bibliography{IEEEabrv,reference_venues,references}

\begin{thebibliography}{10}
\providecommand{\url}[1]{#1}
\csname url@samestyle\endcsname
\providecommand{\newblock}{\relax}
\providecommand{\bibinfo}[2]{#2}
\providecommand{\BIBentrySTDinterwordspacing}{\spaceskip=0pt\relax}
\providecommand{\BIBentryALTinterwordstretchfactor}{4}
\providecommand{\BIBentryALTinterwordspacing}{\spaceskip=\fontdimen2\font plus
\BIBentryALTinterwordstretchfactor\fontdimen3\font minus
  \fontdimen4\font\relax}
\providecommand{\BIBforeignlanguage}[2]{{%
\expandafter\ifx\csname l@#1\endcsname\relax
\typeout{** WARNING: exmrIEEE.bst: No hyphenation pattern has been}%
\typeout{** loaded for the language `#1'. Using the pattern for}%
\typeout{** the default language instead.}%
\else
\language=\csname l@#1\endcsname
\fi
#2}}
\providecommand{\BIBdecl}{\relax}
\BIBdecl

\bibitem{shin2024wham}
S.~Shin, J.~Kim, E.~Halilaj, and M.~J. Black, ``{WHAM}: Reconstructing
  world-grounded humans with accurate {3D} motion,'' in \emph{Proc. IEEE/CVF
  Conf. Comput. Vis. Pattern Recognit.}, 2024, pp. 2070--2080.

\bibitem{wang2024tram}
Y.~Wang, Z.~Wang, L.~Liu, and K.~Daniilidis, ``{TRAM}: Global trajectory and
  motion of {3D} humans from in-the-wild videos,'' in \emph{Proc. Eur. Conf.
  Comput. Vis. (2024)}, 2025, pp. 467--487.

\bibitem{shen2024gvhmr}
Z.~Shen \emph{et~al.}, ``World-grounded human motion recovery via gravity-view
  coordinates,'' in \emph{Proc. SIGGRAPH Asia Conf. Papers}, 2024, pp. 1--11.

\bibitem{wang2025prompthmr}
Y.~Wang, Y.~Sun, P.~Patel, K.~Daniilidis, M.~J. Black, and M.~Kocabas,
  ``{PromptHMR}: Promptable human mesh recovery,'' in \emph{Proc. IEEE/CVF
  Conf. Comput. Vis. Pattern Recognit.}, 2025, pp. 1148--1159.

\bibitem{li2025genmo}
J.~Li \emph{et~al.}, ``{GENMO}: A {GEN}eralist model for human {MO}tion,'' in
  \emph{Proc. IEEE Int. Conf. Comput. Vis.}, 2025, pp. 11\,766--11\,776.

\bibitem{wang2026duomo}
Y.~Wang \emph{et~al.}, ``{DuoMo}: Dual motion diffusion for world-space human
  reconstruction,'' in \emph{Proc. IEEE/CVF Conf. Comput. Vis. Pattern
  Recognit.}, 2026, pp. 42\,777--42\,788.

\bibitem{cai2023smplerx}
Z.~Cai \emph{et~al.}, ``{SMPLer-X}: Scaling up expressive human pose and shape
  estimation,'' in \emph{Proc. Adv. Neural Inf. Process. Syst.}, 2023, pp.
  11\,454--11\,468.

\bibitem{yin2025smplest}
W.~Yin \emph{et~al.}, ``{SMPLest-X}: Ultimate scaling for expressive human pose
  and shape estimation,'' \emph{{IEEE} Trans. Pattern Anal. Mach. Intell.},
  vol.~48, no.~2, pp. 1778--1794, Feb. 2026.

\bibitem{yang2026sam3dbody}
X.~Yang \emph{et~al.}, ``{SAM 3D Body}: Robust full-body human mesh recovery,''
  in \emph{Proc. IEEE/CVF Conf. Comput. Vis. Pattern Recognit.}, 2026, pp.
  7209--7219.

\bibitem{rong2021frankmocap}
Y.~Rong, T.~Shiratori, and H.~Joo, ``{FrankMocap}: A monocular {3D} whole-body
  pose estimation system via regression and integration,'' in \emph{Proc. IEEE
  Int. Conf. Comput. Vis. Workshops}, 2021, pp. 1749--1759.

\bibitem{pymafx2023}
H.~Zhang \emph{et~al.}, ``{PyMAF-X}: Towards well-aligned full-body model
  regression from monocular images,'' \emph{{IEEE} Trans. Pattern Anal. Mach.
  Intell.}, vol.~45, no.~10, pp. 12\,287--12\,303, Oct. 2023.

\bibitem{moon2026h4wpp}
G.~Moon, ``Enhancing hands in {3D} whole-body pose estimation with conditional
  hands modulator,'' in \emph{Proc. IEEE/CVF Conf. Comput. Vis. Pattern
  Recognit.}, 2026, pp. 8891--8900.

\bibitem{Duran_2024_WACV}
E.~Duran, M.~Kocabas, V.~Choutas, Z.~Fan, and M.~J. Black, ``{HMP}: Hand motion
  priors for pose and shape estimation from video,'' in \emph{Proc. IEEE Winter
  Conf. Appl. Comput. Vis.}, 2024, pp. 6341--6351.

\bibitem{zhang2025hawor}
J.~Zhang, J.~Deng, C.~Ma, and R.~A. Potamias, ``{HaWoR}: World-space hand
  motion reconstruction from egocentric videos,'' in \emph{Proc. IEEE/CVF Conf.
  Comput. Vis. Pattern Recognit.}, 2025, pp. 1805--1815.

\bibitem{Ismayilzada_2026_CVPR}
E.~Ismayilzada, Y.~Zhang, and Z.~Cui, ``{PAD-Hand}: Physics-aware diffusion for
  hand motion recovery,'' in \emph{Proc. IEEE/CVF Conf. Comput. Vis. Pattern
  Recognit.}, 2026, pp. 28\,358--28\,368.

\bibitem{yin2024whac}
W.~Yin \emph{et~al.}, ``{WHAC}: World-grounded humans and cameras,'' in
  \emph{Proc. Eur. Conf. Comput. Vis. (2024)}, 2025, pp. 20--37.

\bibitem{Liu_2025_CVPR}
K.~Liu \emph{et~al.}, ``Motions as queries: One-stage multi-person holistic
  human motion capture,'' in \emph{Proc. IEEE/CVF Conf. Comput. Vis. Pattern
  Recognit.}, 2025, pp. 17\,529--17\,539.

\bibitem{chen2026human3r}
Y.~Chen, X.~Chen, Y.~Xue, A.~Chen, Y.~Xiu, and G.~Pons-Moll, ``{Human3R}:
  Everyone everywhere all at once,'' in \emph{Proc. Int. Conf. Learn.
  Representations}, 2026, pp. 42\,584--42\,613.

\bibitem{ExPose:2020}
V.~Choutas, G.~Pavlakos, T.~Bolkart, D.~Tzionas, and M.~J. Black, ``Monocular
  expressive body regression through body-driven attention,'' in \emph{Proc.
  Eur. Conf. Comput. Vis.}, 2020, pp. 20--40.

\bibitem{shen2026dancehmr}
W.~Shen, M.~Zhou, H.~Zhang, S.~Bian, Y.~Xu, and Y.~Zhang, ``{DanceHMR}:
  Hand-aware whole-body human mesh recovery from monocular videos,''
  arXiv:2605.18102, May 2026.

\bibitem{wehrbein26footmr}
T.~Wehrbein and B.~Rosenhahn, ``Improving {3D} foot motion reconstruction in
  markerless monocular human motion capture,'' in \emph{Proc. Int. Conf. 3D
  Vis.}, 2026, pp. 871--881.

\bibitem{carreira2016ief}
J.~Carreira, P.~Agrawal, K.~Fragkiadaki, and J.~Malik, ``Human pose estimation
  with iterative error feedback,'' in \emph{Proc. IEEE/CVF Conf. Comput. Vis.
  Pattern Recognit.}, 2016, pp. 4733--4742.

\bibitem{kanazawa18hmr}
A.~Kanazawa, M.~J. Black, D.~W. Jacobs, and J.~Malik, ``End-to-end recovery of
  human shape and pose,'' in \emph{Proc. IEEE/CVF Conf. Comput. Vis. Pattern
  Recognit.}, 2018, pp. 7122--7131.

\bibitem{luo2020meva}
Z.~Luo, S.~A. Golestaneh, and K.~M. Kitani, ``{3D} human motion estimation via
  motion compression and refinement,'' in \emph{Proc. Asian Conf. Comput. Vis.
  (2020)}, 2021, pp. 324--340.

\bibitem{li2021localrecurrent}
R.~Li, S.~Karanam, R.~Li, T.~Chen, B.~Bhanu, and Z.~Wu, ``Learning local
  recurrent models for human mesh recovery,'' in \emph{Proc. Int. Conf. 3D
  Vis.}, 2021, pp. 555--564.

\bibitem{pavlakos2019smplx}
G.~Pavlakos \emph{et~al.}, ``Expressive body capture: {3D} hands, face, and
  body from a single image,'' in \emph{Proc. IEEE/CVF Conf. Comput. Vis.
  Pattern Recognit.}, 2019, pp. 10\,967--10\,977.

\bibitem{pavlakos2024hamer}
G.~Pavlakos, D.~Shan, I.~Radosavovic, A.~Kanazawa, D.~Fouhey, and J.~Malik,
  ``Reconstructing hands in {3D} with transformers,'' in \emph{Proc. IEEE/CVF
  Conf. Comput. Vis. Pattern Recognit.}, 2024, pp. 9826--9836.

\bibitem{tesch2025bedlam2}
J.~Tesch \emph{et~al.}, ``{BEDLAM2.0}: Synthetic humans and cameras in
  motion,'' in \emph{Proc. Adv. Neural Inf. Process. Syst.}, 2025, pp.
  54\,350--54\,382.

\bibitem{fan2023arctic}
Z.~Fan \emph{et~al.}, ``{ARCTIC}: A dataset for dexterous bimanual hand-object
  manipulation,'' in \emph{Proc. IEEE/CVF Conf. Comput. Vis. Pattern
  Recognit.}, 2023, pp. 12\,943--12\,954.

\bibitem{tripathi2023moyo}
S.~Tripathi, L.~M\"uller, C.-H.~P. Huang, O.~Taheri, M.~J. Black, and
  D.~Tzionas, ``{3D} human pose estimation via intuitive physics,'' in
  \emph{Proc. IEEE/CVF Conf. Comput. Vis. Pattern Recognit.}, 2023, pp.
  4713--4725.

\bibitem{huang2022rich}
C.-H.~P. Huang \emph{et~al.}, ``Capturing and inferring dense full-body
  human-scene contact,'' in \emph{Proc. IEEE/CVF Conf. Comput. Vis. Pattern
  Recognit.}, 2022, pp. 13\,264--13\,275.

\bibitem{loper2015smpl}
M.~Loper, N.~Mahmood, J.~Romero, G.~Pons-Moll, and M.~J. Black, ``{SMPL}: A
  skinned multi-person linear model,'' \emph{ACM Trans. Graph.}, vol.~34,
  no.~6, pp. 1--16, Nov. 2015, art.~no.~248.

\bibitem{bogo16eccv}
F.~Bogo, A.~Kanazawa, C.~Lassner, P.~Gehler, J.~Romero, and M.~J. Black, ``Keep
  it {SMPL}: Automatic estimation of {3D} human pose and shape from a single
  image,'' in \emph{Proc. Eur. Conf. Comput. Vis.}, 2016, pp. 561--578.

\bibitem{li2022cliff}
Z.~Li, J.~Liu, Z.~Zhang, S.~Xu, and Y.~Yan, ``{CLIFF}: Carrying location
  information in full frames into human pose and shape estimation,'' in
  \emph{Proc. Eur. Conf. Comput. Vis.}, 2022, pp. 590--606.

\bibitem{goel2023humans}
S.~Goel, G.~Pavlakos, J.~Rajasegaran, A.~Kanazawa, and J.~Malik, ``Humans in
  {4D}: Reconstructing and tracking humans with transformers,'' in \emph{Proc.
  IEEE Int. Conf. Comput. Vis.}, 2023, pp. 14\,737--14\,748.

\bibitem{sun2024aios}
Q.~Sun \emph{et~al.}, ``{AiOS}: All-in-one-stage expressive human pose and
  shape estimation,'' in \emph{Proc. IEEE/CVF Conf. Comput. Vis. Pattern
  Recognit.}, 2024, pp. 1834--1843.

\bibitem{ferguson2025mhr}
A.~Ferguson \emph{et~al.}, ``{MHR}: Momentum human rig,'' arXiv:2511.15586,
  Nov. 2025.

\bibitem{simoni2026dinov}
\BIBentryALTinterwordspacing
O.~Sim{\'e}oni \emph{et~al.}, ``{DINOv3},'' \emph{Trans. Mach. Learn. Res.},
  May 2026. [Online]. Available:
  \url{https://openreview.net/forum?id=2NlGyqNjns}
\BIBentrySTDinterwordspacing

\bibitem{romero2017mano}
J.~Romero, D.~Tzionas, and M.~J. Black, ``Embodied hands: Modeling and
  capturing hands and bodies together,'' \emph{ACM Trans. Graph.}, vol.~36,
  no.~6, pp. 1--17, Nov. 2017, art.~no.~245.

\bibitem{Potamias_2025_CVPR}
R.~A. Potamias, J.~Zhang, J.~Deng, and S.~Zafeiriou, ``{WiLoR}: End-to-end {3D}
  hand localization and reconstruction in-the-wild,'' in \emph{Proc. IEEE/CVF
  Conf. Comput. Vis. Pattern Recognit.}, 2025, pp. 12\,242--12\,254.

\bibitem{feng2021pixie}
Y.~Feng, V.~Choutas, T.~Bolkart, D.~Tzionas, and M.~J. Black, ``Collaborative
  regression of expressive bodies using moderation,'' in \emph{Proc. Int. Conf.
  3D Vis.}, 2021, pp. 792--804.

\bibitem{Moon_2022_CVPR}
G.~Moon, H.~Choi, and K.~M. Lee, ``Accurate {3D} hand pose estimation for
  whole-body {3D} human mesh estimation,'' in \emph{Proc. IEEE/CVF Conf.
  Comput. Vis. Pattern Recognit. Workshops}, 2022, pp. 2307--2316.

\bibitem{shen2024hmradapter}
W.~Shen \emph{et~al.}, ``{HMR-Adapter}: A lightweight adapter with dual-path
  cross augmentation for expressive human mesh recovery,'' in \emph{Proc. ACM
  Int. Conf. Multimedia}, 2024, pp. 6093--6102.

\bibitem{kanazawa2019hmmr}
A.~Kanazawa, J.~Y. Zhang, P.~Felsen, and J.~Malik, ``Learning {3D} human
  dynamics from video,'' in \emph{Proc. IEEE/CVF Conf. Comput. Vis. Pattern
  Recognit.}, 2019, pp. 5607--5616.

\bibitem{kocabas2019vibe}
M.~Kocabas, N.~Athanasiou, and M.~J. Black, ``{VIBE}: Video inference for human
  body pose and shape estimation,'' in \emph{Proc. IEEE/CVF Conf. Comput. Vis.
  Pattern Recognit.}, 2020, pp. 5252--5262.

\bibitem{choi2020beyond}
H.~Choi, G.~Moon, J.~Y. Chang, and K.~M. Lee, ``Beyond static features for
  temporally consistent {3D} human pose and shape from a video,'' in
  \emph{Proc. IEEE/CVF Conf. Comput. Vis. Pattern Recognit.}, 2021, pp.
  1964--1973.

\bibitem{WeiLin2022mpsnet}
W.-L. Wei, J.-C. Lin, T.-L. Liu, and H.-Y.~M. Liao, ``Capturing humans in
  motion: Temporal-attentive {3D} human pose and shape estimation from
  monocular video,'' in \emph{Proc. IEEE/CVF Conf. Comput. Vis. Pattern
  Recognit.}, 2022, pp. 13\,201--13\,210.

\bibitem{shen2023global}
X.~Shen, Z.~Yang, X.~Wang, J.~Ma, C.~Zhou, and Y.~Yang, ``Global-to-local
  modeling for video-based {3D} human pose and shape estimation,'' in
  \emph{Proc. IEEE/CVF Conf. Comput. Vis. Pattern Recognit.}, 2023, pp.
  8887--8896.

\bibitem{yuan2022glamr}
Y.~Yuan, U.~Iqbal, P.~Molchanov, K.~Kitani, and J.~Kautz, ``{GLAMR}: Global
  occlusion-aware human mesh recovery with dynamic cameras,'' in \emph{Proc.
  IEEE/CVF Conf. Comput. Vis. Pattern Recognit.}, 2022, pp. 11\,028--11\,039.

\bibitem{ye2023slahmr}
V.~Ye, G.~Pavlakos, J.~Malik, and A.~Kanazawa, ``Decoupling human and camera
  motion from videos in the wild,'' in \emph{Proc. IEEE/CVF Conf. Comput. Vis.
  Pattern Recognit.}, 2023, pp. 21\,222--21\,232.

\bibitem{xiang2019monocular}
D.~Xiang, H.~Joo, and Y.~Sheikh, ``Monocular total capture: Posing face, body,
  and hands in the wild,'' in \emph{Proc. IEEE/CVF Conf. Comput. Vis. Pattern
  Recognit.}, 2019, pp. 10\,957--10\,966.

\bibitem{cao2019openpose}
Z.~Cao, G.~Hidalgo, T.~Simon, S.-E. Wei, and Y.~Sheikh, ``{OpenPose}: Realtime
  multi-person {2D} pose estimation using part affinity fields,'' \emph{{IEEE}
  Trans. Pattern Anal. Mach. Intell.}, vol.~43, no.~1, pp. 172--186, Jan. 2021.

\bibitem{jin2020whole}
S.~Jin \emph{et~al.}, ``Whole-body human pose estimation in the wild,'' in
  \emph{Proc. Eur. Conf. Comput. Vis.}, 2020, pp. 196--214.

\bibitem{xu2022vitpose+}
Y.~Xu, J.~Zhang, Q.~Zhang, and D.~Tao, ``{ViTPose++}: Vision transformer for
  generic body pose estimation,'' \emph{{IEEE} Trans. Pattern Anal. Mach.
  Intell.}, vol.~46, no.~2, pp. 1212--1230, Feb. 2024.

\bibitem{jiang2024rtmw}
T.~Jiang, X.~Xie, and Y.~Li, ``{RTMW}: Real-time multi-person {2D} and {3D}
  whole-body pose estimation,'' arXiv:2407.08634, Jul. 2024.

\bibitem{khirodkar2024sapiens}
R.~Khirodkar \emph{et~al.}, ``{Sapiens}: Foundation for human vision models,''
  in \emph{Proc. Eur. Conf. Comput. Vis. (2024)}, 2025, pp. 206--228.

\bibitem{khirodkar2026sapiens2}
R.~Khirodkar, H.~Wen, J.~Martinez, Y.~Dong, Z.~Su, and S.~Saito,
  ``{Sapiens2},'' in \emph{Proc. Int. Conf. Learn. Representations}, 2026, pp.
  58\,484--58\,507.

\bibitem{zhuo2023foot}
L.~Zhuo, J.~Cao, Q.~Wang, B.~Zhang, and L.~Bo, ``Towards stable human pose
  estimation via cross-view fusion and foot stabilization,'' in \emph{Proc.
  IEEE/CVF Conf. Comput. Vis. Pattern Recognit.}, 2023, pp. 650--659.

\bibitem{zou2020footskate}
Y.~Zou, J.~Yang, D.~Ceylan, J.~Zhang, F.~Perazzi, and J.-B. Huang, ``Reducing
  footskate in human motion reconstruction with ground contact constraints,''
  in \emph{Proc. IEEE Winter Conf. Appl. Comput. Vis.}, 2020, pp. 448--457.

\bibitem{rempe2020contact}
D.~Rempe, L.~J. Guibas, A.~Hertzmann, B.~Russell, R.~Villegas, and J.~Yang,
  ``Contact and human dynamics from monocular video,'' in \emph{Proc. Eur.
  Conf. Comput. Vis.}, 2020, pp. 71--87.

\bibitem{shimada2020physcap}
S.~Shimada, V.~Golyanik, W.~Xu, and C.~Theobalt, ``{PhysCap}: Physically
  plausible monocular {3D} motion capture in real time,'' \emph{ACM Trans.
  Graph.}, vol.~39, no.~6, pp. 1--16, Dec. 2020, art.~no.~235.

\bibitem{rempe2021humor}
D.~Rempe, T.~Birdal, A.~Hertzmann, J.~Yang, S.~Sridhar, and L.~J. Guibas,
  ``{HuMoR}: {3D} human motion model for robust pose estimation,'' in
  \emph{Proc. IEEE Int. Conf. Comput. Vis.}, 2021, pp. 11\,468--11\,479.

\bibitem{zhang21pymaf}
H.~Zhang \emph{et~al.}, ``{PyMAF}: {3D} human pose and shape regression with
  pyramidal mesh alignment feedback loop,'' in \emph{Proc. IEEE Int. Conf.
  Comput. Vis.}, 2021, pp. 11\,426--11\,436.

\bibitem{wang2023refit}
Y.~Wang and K.~Daniilidis, ``{ReFit}: Recurrent fitting network for {3D} human
  recovery,'' in \emph{Proc. IEEE Int. Conf. Comput. Vis.}, 2023, pp.
  14\,598--14\,608.

\bibitem{fieraru2018refine}
M.~Fieraru, A.~Khoreva, L.~Pishchulin, and B.~Schiele, ``Learning to refine
  human pose estimation,'' in \emph{Proc. IEEE/CVF Conf. Comput. Vis. Pattern
  Recognit. Workshops}, 2018, pp. 205--214.

\bibitem{moon2019posefix}
G.~Moon, J.~Y. Chang, and K.~M. Lee, ``{PoseFix}: Model-agnostic general human
  pose refinement network,'' in \emph{Proc. IEEE/CVF Conf. Comput. Vis. Pattern
  Recognit.}, 2019, pp. 7765--7773.

\bibitem{wehrbein2023iccvw}
T.~Wehrbein, B.~Rosenhahn, I.~Matthews, and C.~Stoll, ``Personalized {3D} human
  pose and shape refinement,'' in \emph{Proc. IEEE Int. Conf. Comput. Vis.
  Workshops}, 2023, pp. 4191--4201.

\bibitem{baradel2023posebert}
F.~Baradel, R.~Br{\'e}gier, T.~Groueix, P.~Weinzaepfel, Y.~Kalantidis, and
  G.~Rogez, ``{PoseBERT}: A generic transformer module for temporal {3D} human
  modeling,'' \emph{{IEEE} Trans. Pattern Anal. Mach. Intell.}, vol.~45,
  no.~11, pp. 12\,798--12\,815, Nov. 2023.

\bibitem{zeng2022smoothnet}
A.~Zeng, L.~Yang, X.~Ju, J.~Li, J.~Wang, and Q.~Xu, ``{SmoothNet}: A
  plug-and-play network for refining human poses in videos,'' in \emph{Proc.
  Eur. Conf. Comput. Vis.}, 2022, pp. 625--642.

\bibitem{wang2024synsp}
T.~Wang \emph{et~al.}, ``{SynSP}: Synergy of smoothness and precision in pose
  sequences refinement,'' in \emph{Proc. IEEE/CVF Conf. Comput. Vis. Pattern
  Recognit.}, 2024, pp. 1824--1833.

\bibitem{lee2021partdynamics}
G.-H. Lee and S.-W. Lee, ``Uncertainty-aware human mesh recovery from video by
  learning part-based {3D} dynamics,'' in \emph{Proc. IEEE Int. Conf. Comput.
  Vis.}, 2021, pp. 12\,355--12\,364.

\bibitem{zhou19rot6d}
Y.~Zhou, C.~Barnes, J.~Lu, J.~Yang, and H.~Li, ``On the continuity of rotation
  representations in neural networks,'' in \emph{Proc. IEEE/CVF Conf. Comput.
  Vis. Pattern Recognit.}, 2019, pp. 5738--5746.

\bibitem{lin2014coco}
T.-Y. Lin \emph{et~al.}, ``{Microsoft COCO}: Common objects in context,'' in
  \emph{Proc. Eur. Conf. Comput. Vis.}, 2014, pp. 740--755.

\bibitem{su2024rope}
J.~Su, M.~Ahmed, Y.~Lu, S.~Pan, W.~Bo, and Y.~Liu, ``{RoFormer}: Enhanced
  transformer with rotary position embedding,'' \emph{Neurocomputing}, vol.
  568, Feb. 2024, art.~no.~127063.

\bibitem{shin2023bio}
S.~Shin, Z.~Li, and E.~Halilaj, ``Markerless motion tracking with noisy video
  and {IMU} data,'' \emph{{IEEE} Trans. Biomed. Eng.}, vol.~70, no.~11, pp.
  3082--3092, Nov. 2023.

\bibitem{ionescu2014h36m}
C.~Ionescu, D.~Papava, V.~Olaru, and C.~Sminchisescu, ``{Human3.6M}: Large
  scale datasets and predictive methods for {3D} human sensing in natural
  environments,'' \emph{{IEEE} Trans. Pattern Anal. Mach. Intell.}, vol.~36,
  no.~7, pp. 1325--1339, Jul. 2014.

\bibitem{marcard2018eccv}
T.~von Marcard, R.~Henschel, M.~J. Black, B.~Rosenhahn, and G.~Pons-Moll,
  ``Recovering accurate {3D} human pose in the wild using {IMUs} and a moving
  camera,'' in \emph{Proc. Eur. Conf. Comput. Vis.}, 2018, pp. 614--631.

\bibitem{kaufmann2023emdb}
M.~Kaufmann \emph{et~al.}, ``{EMDB}: The electromagnetic database of global
  {3D} human pose and shape in the wild,'' in \emph{Proc. IEEE Int. Conf.
  Comput. Vis.}, 2023, pp. 14\,586--14\,597.

\bibitem{rhodin2018cvpr}
H.~Rhodin \emph{et~al.}, ``Learning monocular {3D} human pose estimation from
  multi-view images,'' in \emph{Proc. IEEE/CVF Conf. Comput. Vis. Pattern
  Recognit.}, 2018, pp. 8437--8446.

\bibitem{black2023bedlam}
M.~J. Black, P.~Patel, J.~Tesch, and J.~Yang, ``{BEDLAM}: A synthetic dataset
  of bodies exhibiting detailed lifelike animated motion,'' in \emph{Proc.
  IEEE/CVF Conf. Comput. Vis. Pattern Recognit.}, 2023, pp. 8726--8737.

\bibitem{mahmood2019amass}
N.~Mahmood, N.~Ghorbani, N.~F. Troje, G.~Pons-Moll, and M.~J. Black, ``{AMASS}:
  Archive of motion capture as surface shapes,'' in \emph{Proc. IEEE Int. Conf.
  Comput. Vis.}, 2019, pp. 5441--5450.

\bibitem{loshchilov2019adamw}
\BIBentryALTinterwordspacing
I.~Loshchilov and F.~Hutter, ``Decoupled weight decay regularization,'' in
  \emph{Proc. Int. Conf. Learn. Representations}, 2019. [Online]. Available:
  \url{https://openreview.net/forum?id=Bkg6RiCqY7}
\BIBentrySTDinterwordspacing

\bibitem{dwivedi2024tokenhmr}
S.~K. Dwivedi, Y.~Sun, P.~Patel, Y.~Feng, and M.~J. Black, ``{TokenHMR}:
  Advancing human mesh recovery with a tokenized pose representation,'' in
  \emph{Proc. IEEE/CVF Conf. Comput. Vis. Pattern Recognit.}, 2024, pp.
  1323--1333.

\bibitem{patel2024camerahmr}
P.~Patel and M.~J. Black, ``{CameraHMR}: Aligning people with perspective,'' in
  \emph{Proc. Int. Conf. 3D Vis.}, 2025, pp. 1562--1571.

\bibitem{sarandi2024nlf}
I.~S\'ar\'andi and G.~Pons-Moll, ``Neural localizer fields for continuous {3D}
  human pose and shape estimation,'' in \emph{Proc. Adv. Neural Inf. Process.
  Syst.}, 2024, pp. 140\,032--140\,065.

\bibitem{xu22vitpose}
Y.~Xu, J.~Zhang, Q.~Zhang, and D.~Tao, ``{ViTPose}: Simple vision transformer
  baselines for human pose estimation,'' in \emph{Proc. Adv. Neural Inf.
  Process. Syst.}, 2022, pp. 38\,571--38\,584.

\end{thebibliography}

\begin{IEEEbiography}
[{\includegraphics[width=1in,height=1.25in,clip,keepaspectratio]{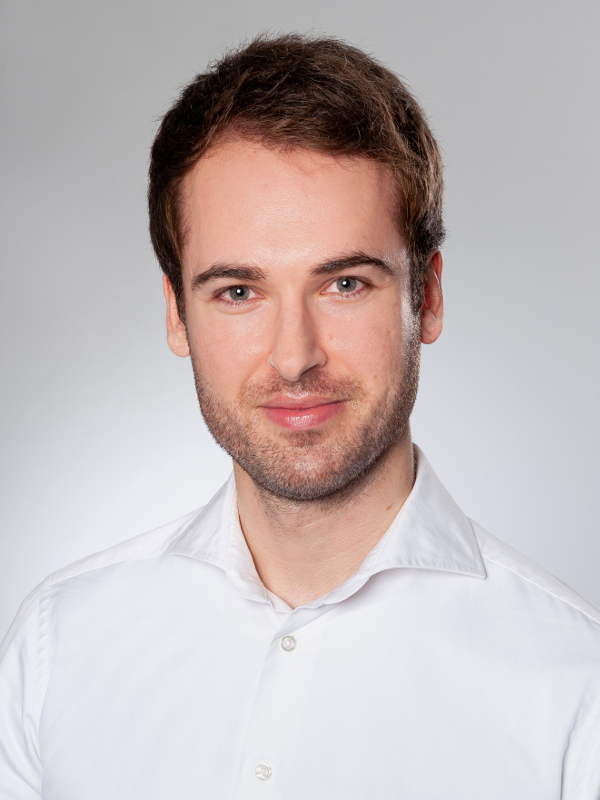}}]
{Tom Wehrbein}
received the B.Sc.\ and M.Sc.\ degrees in computer science from Leibniz
University Hannover in 2018 and 2020, respectively. He is currently pursuing
the Ph.D.\ degree at the Institute of Information Processing, Leibniz
University Hannover, under the supervision of Prof.\ Bodo Rosenhahn.
In 2022, he was an intern at Epic Games in Pittsburgh. His research interests
include computer vision, with a focus on human motion capture.
\end{IEEEbiography}

\begin{IEEEbiography}
[{\includegraphics[width=1in,height=1.25in,clip,keepaspectratio]{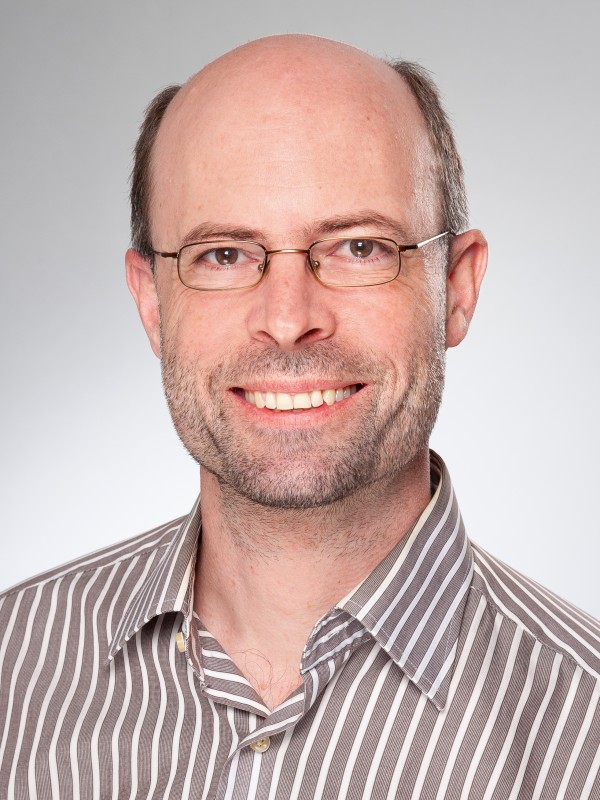}}]
{Bodo Rosenhahn}
received the Dipl.-Inf.\ and Dr.-Ing.\ degrees from the University of Kiel
in 1999 and 2003, respectively. He studied computer science, with medicine
as a minor subject, at the University of Kiel. From October 2003 to October 2005,
he worked as a postdoctoral researcher at the University of Auckland, New Zealand,
funded by a scholarship from the German Research Foundation (DFG).
From November 2005 to August 2008, he worked as a senior researcher
at the Max Planck Institute for Computer Science. Since September 2008,
he has been a full professor at Leibniz University Hannover, heading a group
on automated image interpretation. He has coauthored more than 200 papers,
holds 12 patents, and has organized several workshops and conferences.
His work has received several awards, including the DAGM Prize in 2002,
the Siegfried Werth Prize in 2003, the DAGM Main Prize in 2005 and 2007,
the Olympus Prize in 2007, and the G{\"u}nter Enderle Award (Eurographics) in 2017.
\end{IEEEbiography}

\vfill

\clearpage
\setcounter{subsection}{0}
\setcounter{equation}{0}
\renewcommand{\theHsubsection}{S\arabic{subsection}}
\renewcommand{\theHequation}{S\arabic{equation}}
\renewcommand{\theHtable}{S\arabic{table}}
\renewcommand{\theHfigure}{S\arabic{figure}}

\renewcommand{\thesubsection}{S\arabic{subsection}}
\renewcommand{\thesubsectiondis}{\thesubsection.}
\renewcommand{\theequation}{S\arabic{equation}}
\section*{Supplementary Material}
The supplementary material begins with the neutral SMPL-X shape fitting used
to construct ARCTIC training supervision
(Section~\ref{sec:arctic_shape_fitting}). It then details the HaMeR copy-paste
protocol (Section~\ref{sec:hamer_copy_paste_protocol}) and analyzes inference time and the
cost--accuracy tradeoff of alternative 2D keypoint detectors
(Section~\ref{sec:inference_time_analysis}). The next subsection revisits the
inherited FootMR design choices within \modelname\
(Section~\ref{sec:footmr_design_analysis}). Finally, we present qualitative
comparisons with SAM3D-Body
(Section~\ref{sec:supplement_qualitative}).
The accompanying supplementary video illustrates reconstructed full-body motion over
time.

\setcounter{table}{0}
\renewcommand{\thetable}{S\arabic{table}}
\setcounter{figure}{0}
\renewcommand{\thefigure}{S\arabic{figure}}

\subsection{ARCTIC Training-Subject Shape Fitting}
\label{sec:arctic_shape_fitting}
ARCTIC~\cite{fan2023arctic} does not provide SMPL-X shape coefficients for its
body annotations. Instead, it represents each subject by a personalized
template and a male or female label, together with per-frame pose annotations.
\modelname\ uses the neutral SMPL-X model for every input, without assuming
or estimating gender labels at inference. We therefore fit neutral SMPL-X
shape coefficients to ARCTIC's ground-truth meshes to obtain compatible
training supervision.

For each of the eight ARCTIC training subjects, we fit one
10-dimensional neutral shape vector using $1{,}000$ frames sampled uniformly
across the subject's annotations. The personalized gender-specific and neutral
SMPL-X meshes are constructed with identical annotated body and hand poses,
and all pose parameters remain fixed during fitting. We optimize the neutral shape
vector with L-BFGS, minimizing the mean squared coordinate difference between
corresponding vertices over the full $10{,}475$-vertex mesh.

Across all annotated frames from the training subjects, the mean Euclidean
distance between corresponding vertices of the fitted neutral and ground-truth
meshes is $6.24\ \mathrm{mm}$, with subject means ranging from $3.08$ to $9.86\ \mathrm{mm}$.
The fitted shapes are only used to construct neutral-model training supervision.
Evaluation directly uses ARCTIC's personalized gender-specific
ground-truth meshes.

\begin{table}[t]
\scriptsize
\centering
\caption{\textbf{Runtime breakdown of \modelname\ with alternative 2D detectors.}
    Times are in milliseconds per video frame for one reconstructed person,
    measured over complete passes through the video after warm-up.
    Sapiens2-1B uses the default $1024\times768$-pixel input resolution.
    Totals are sums of stage medians, not measured end-to-end times.
}\label{table:inference_runtime}
\begin{tabular}{@{}lc@{}}
    \toprule
    Stage & Time ($\mathrm{ms/frame}$) \\
    \midrule
    Person detection and tracking & 3.02 \\
    SAM3D-Body feature extraction & 25.33 \\
    HaMeR feature extraction & 6.34 \\
    2D keypoint detection: Sapiens2-1B & 118.62 \\
    2D keypoint detection: RTMW-x & 2.11 \\
    \midrule
    Initial estimator & 0.0119 \\
    FootMR module & 0.0029 \\
    HandMR module & 0.0047 \\
    \midrule
    Total with Sapiens2-1B & 153.33 \\
    Total with RTMW-x & 36.81 \\
    \bottomrule
\end{tabular}
\end{table}

\subsection{HaMeR Copy-Paste Protocol}
\label{sec:hamer_copy_paste_protocol}
The \modelname\ + HaMeR copy-paste variant in
Table~\ref{table:hand_integration_evidence_ablation} starts from the final
\modelname\ prediction, including HandMR refinement. It uses the same trained
\modelname\ checkpoint, fixed HaMeR checkpoint, and hand crops, and replaces
only the wrist and finger rotations with HaMeR outputs from the same forward
pass.
Because HaMeR uses a canonical right-hand convention, we first transform the
predictions from mirrored left-hand crops back to the left-hand convention.
We then insert its parent-relative finger rotations and convert its crop-frame
global wrist rotation to the elbow-relative rotation required by SMPL-X:
\begin{equation}
    \mathbf{R}_w^{(l)}=
    \left(\bar{\mathbf{R}}_e^{(l)}\right)^{\!\top}
    \bar{\mathbf{R}}_{w,\mathrm{HaMeR}}^{(l)}.
\end{equation}
Here, \(\bar{\mathbf{R}}_{w,\mathrm{HaMeR}}^{(l)}\) denotes HaMeR's global
wrist rotation after the left-hand convention correction, and the global elbow
rotation \(\bar{\mathbf{R}}_e^{(l)}\) is retained from \modelname.

\begin{table*}[t]
\scriptsize
\centering
\caption{\textbf{Effect of the 2D keypoint detector on reconstruction accuracy and runtime.}
    All configurations use the same trained \modelname\ checkpoint.
    Accuracy is evaluated on the benchmark datasets with person and hand boxes
    held fixed across configurations, following the dataset-specific protocol in
    Section~\ref{sec:experimental_protocol}. Only the 23 body-and-foot keypoint
    inputs change. Detector and summed stage costs are measured on the
    demonstration video following the timing protocol in Table~\ref{table:inference_runtime}.
    PVE and MPJPE are in millimeters, AJAE is in degrees, and N-FKE\textsubscript{2d} is in normalized pixel space.
}\label{table:detector_speed_accuracy}
\setlength{\tabcolsep}{4.8pt}
\begin{tabular}{@{}lcccccccccccc@{}}
    \toprule
    & \multicolumn{2}{c}{Time ($\mathrm{ms/frame}$)} & ARCTIC &
    \multicolumn{3}{c}{MOYO} & \multicolumn{3}{c}{RICH} & MOOF & 3DPW & EMDB-1 \\
    \cmidrule(lr){2-3}\cmidrule(lr){4-4}\cmidrule(lr){5-7}\cmidrule(lr){8-10}\cmidrule(lr){11-11}\cmidrule(lr){12-12}\cmidrule(lr){13-13}
    2D detector & Detector & Summed stages & Hand PVE &
    All PVE & Hand PVE & AJAE & All PVE & Hand PVE & AJAE & N-FKE\textsubscript{2d} & MPJPE & MPJPE \\
    \midrule
    Sapiens2-1B ($1024\times768$) & 118.62 & 153.33 & 14.7 & 64.2 & 26.5 & 20.0 & 58.4 & 21.9 & 12.8 & 0.67 & 51.8 & 64.1 \\
    Sapiens2-1B ($576\times432$) & 33.06 & 67.77 & 14.7 & 64.2 & 26.5 & 19.8 & 58.5 & 21.9 & 12.9 & 0.67 & 51.7 & 64.1 \\
    Sapiens2-1B ($448\times336$) & 19.77 & 54.47 & 14.7 & 64.1 & 26.5 & 19.9 & 58.6 & 21.9 & 13.1 & 0.68 & 51.7 & 64.1 \\
    RTMW-x & 2.11 & 36.81 & 14.7 & 65.1 & 26.6 & 21.8 & 58.9 & 22.0 & 13.7 & 0.69 & 51.8 & 64.8 \\
    \bottomrule
\end{tabular}
\end{table*}

\begin{table*}[t]
\scriptsize
\centering
\caption{\textbf{Implementation settings for inference-time measurements.}
    Spatial input sizes are height $\times$ width in pixels.
    Batch sizes count person crops except where stated.
    Worker counts refer to data loader workers. A dash denotes a worker count
    that is not applicable or no additional implementation notes.
    Frame prefetching supplies decoded video frames from a background worker.
}\label{table:inference_settings}
\begin{tabular}{@{}llllll@{}}
    \toprule
    Stage & Input & Inference precision & Batch size & Workers & Implementation notes \\
    \midrule
    Detection/tracking & $384\times640$ & FP16 & 8 & --- & YOLO11m, ByteTrack; frame prefetching \\
    SAM3D-Body & $512\times512$ & BF16 backbone & 32 & 8 & --- \\
    HaMeR & $256\times192$ & FP16 autocast & 32 hand crops & 8 & --- \\
    Sapiens2-1B & $1024\times768$ & FP16 & 1 & 2 & FP32 preprocessing; GPU heatmap decoding \\
    Sapiens2-1B & $576\times432$ & FP16 & 2 & 2 & FP32 preprocessing; GPU heatmap decoding \\
    Sapiens2-1B & $448\times336$ & FP16 & 4 & 2 & FP32 preprocessing; GPU heatmap decoding \\
    RTMW-x & $384\times288$ & FP16 & 32 & 8 & ONNX Runtime CUDA (FP32 input/output) \\
    Initial estimator, FootMR, HandMR & Full sequence & FP32 & 1 sequence & --- & PyTorch inference mode \\
    \bottomrule
\end{tabular}
\end{table*}

\subsection{Inference-Time Analysis}
\label{sec:inference_time_analysis}
We analyze the cost of offline, single-person reconstruction and the
cost--accuracy tradeoff of alternative 2D detectors.

\subsubsection{Runtime Measurements}
We measure all stages on the same $3{,}909$-frame, $1920\times1080$-pixel, $30\ \mathrm{frames/s}$
video using an NVIDIA RTX~4090 and Intel Core i9-12900K. After one complete-video warm-up, we time
three complete passes with CUDA synchronization at each interval boundary
and report median wall time per frame. The totals in
Table~\ref{table:inference_runtime} sum unrounded stage medians and exclude
camera-motion estimation and model loading. They are not measured end-to-end
wall times.
Table~\ref{table:inference_settings} records implementation settings.

Feature extraction and keypoint detection dominate the measured cost:
Sapiens2-1B alone accounts for approximately $77\%$ of the default summed
stage cost. FootMR and HandMR take $11.5$ and $18.5\ \mathrm{ms}$ for the complete
sequence, compared with $46.6\ \mathrm{ms}$ for
the initial estimator. Their combined cost is approximately $64\%$ of the
initial-estimator runtime but only $0.021\%$ of the summed stage cost even with
the faster \mbox{RTMW-x} 2D detector~\cite{jiang2024rtmw}.
The estimator and refiner timings include stage input preparation, network
inference, and construction of the refined camera-space SMPL-X parameters.

We also measure the end-to-end runtime of \modelname\ with RTMW-x on the
same video and hardware, from invocation with a video path through saving
camera-space and global SMPL-X motion. The timed interval includes imports,
model construction and loading, JPEG extraction, tracking, keypoint detection,
feature extraction, reconstruction, CPU transfer, and saving.
Stages execute sequentially, and camera-motion estimation is disabled.
Across three independent invocations, the median end-to-end time is
$170.75\ \mathrm{s}$ ($43.68\ \mathrm{ms/frame}$). This includes startup and first-use costs
omitted from the $36.81\ \mathrm{ms/frame}$ sum of warmed stages in
Table~\ref{table:inference_runtime}.

\subsubsection{Detector Cost--Accuracy Tradeoff}
The cost of the default detector motivates testing less expensive configurations.
\modelname\ is trained with perturbed and randomly masked projections of
ground-truth motion rather than outputs from a particular detector
(Section~\ref{sec:training}). We therefore evaluate whether the same checkpoint
can use body-and-foot keypoints from lower-resolution Sapiens2-1B inference
or the much faster RTMW-x without retraining
(Table~\ref{table:detector_speed_accuracy}).
We use a keypoint confidence threshold of $5.5$ for \mbox{RTMW-x} instead of
$0.5$ for Sapiens2-1B, since RTMW-x returns raw unnormalized SimCC
scores.
For runtime comparisons, each configuration uses the batch size listed in
Table~\ref{table:inference_settings}.

The $448\times336$-pixel Sapiens2-1B configuration lowers the summed stage cost from
$153.33$ to $54.47\ \mathrm{ms/frame}$, a factor of $2.81$, with similar reconstruction
accuracy across the reported metrics. RTMW-x reduces the cost to
$36.81\ \mathrm{ms/frame}$, a factor of $4.17$ relative to the
default. Body errors remain close to the default, and wrist-aligned hand PVE
varies by at most $0.1\ \mathrm{mm}$ at the reported precision. MOYO and RICH AJAE increase from $20.0^\circ$
to $21.8^\circ$ and from $12.8^\circ$ to $13.7^\circ$, respectively.
The tested configurations thus show that the body-and-foot keypoint detector
can be replaced without retraining,
with a cost--accuracy tradeoff that is most apparent in foot reconstruction.

\subsection{Design Analysis of the Inherited FootMR Module}
\label{sec:footmr_design_analysis}

\begin{table}[H]
\scriptsize
\centering
\caption{\textbf{Ablation of the FootMR module within \modelname.}
    All \modelname\ configurations are retrained from scratch with otherwise identical settings.
    Top: initial joint-rotation inputs to FootMR, all expressed in the global
    camera frame. When no initial ankle rotation is supplied, FootMR directly
    regresses the global ankle rotation rather than predicting a residual update.
    Bottom: variants of
    the \(\blacktriangleright\)-marked final \modelname\ configuration.
    AJAE is in degrees, N-FKE\textsubscript{2d} in normalized pixel space (lower is better).
}\label{table:ablation_footmr_design}
    \setlength{\tabcolsep}{4pt}
    \begin{tabular}{@{}cccccccc@{}}
        \toprule
        & \multicolumn{4}{c}{Input Joint Rotations} & MOYO & RICH & MOOF \\
        \cmidrule(lr){2-5}\cmidrule(lr){6-8}
        & Ankle & Knee & Hip & Pelvis & AJAE & AJAE & N-FKE\textsubscript{2d} \\
        \midrule
        & \(\times\) & \(\times\) & \(\times\) & \(\times\) & 29.2 & 19.4 & 0.63 \\
        & \(\surd\) & \(\times\) & \(\times\) & \(\times\) & 20.2 & 14.5 & 0.61 \\
        & \(\times\) & \(\surd\) & \(\times\) & \(\times\) & 20.9 & 13.5 & 0.73 \\
        \(\blacktriangleright\) & \(\surd\) & \(\surd\) & \(\times\) & \(\times\) & 20.0 & 12.8 & 0.67 \\
        & \(\surd\) & \(\surd\) & \(\surd\) & \(\times\) & 20.1 & 12.8 & 0.67 \\
        & \(\surd\) & \(\surd\) & \(\surd\) & \(\surd\) & 20.2 & 12.8 & 0.68 \\
        \midrule
        \multicolumn{5}{@{}l}{\(\blacktriangleright\) with knee-relative ankle rotations} & 22.4 & 13.1 & 0.91 \\
        \multicolumn{5}{@{}l}{\(\blacktriangleright\) without global-orientation aug.} & 22.9 & 13.4 & 0.81 \\
        \bottomrule
    \end{tabular}
\end{table}

We revisit three FootMR design choices from our conference
paper~\cite{wehrbein26footmr}: its
joint-rotation inputs, ankle-rotation representation, and global-orientation
augmentation. These ablations test whether they remain beneficial within the
complete \modelname\ system.

\subsubsection{Rotation Inputs and Prediction Formulation}
The top block of Table~\ref{table:ablation_footmr_design} compares FootMR's
joint-rotation inputs. Supplying initial ankle rotations also changes prediction
from direct regression to residual refinement. Both the ankle-only and knee-only
variants achieve lower AJAE on MOYO and RICH than the variant without rotation
inputs. Combining ankle and knee rotations gives the lowest AJAE on both datasets,
reducing MOYO AJAE from
$29.2^\circ$ to $20.0^\circ$ and RICH AJAE from $19.4^\circ$ to $12.8^\circ$.
Adding hip and pelvis rotations yields no further improvement.
These results suggest that the initial ankle estimate provides a useful prior
and that global knee orientations contribute complementary kinematic context
for lifting the 2D foot-keypoint sequence to 3D.

The no-rotation-input and ankle-only variants nevertheless obtain lower MOOF
N-FKE\textsubscript{2d} than the ankle-and-knee configuration
($0.63$ and $0.61$ versus $0.67$). This mismatch shows that accurate 2D foot
alignment alone does not necessarily imply accurate 3D ankle orientation, because
distinct 3D configurations can have similar 2D projections.

\subsubsection{Ankle-Rotation Representation}
Using knee-relative rather than global ankle rotations for both the inputs and
residual updates
increases AJAE from $20.0^\circ$ to $22.4^\circ$ on MOYO and from $12.8^\circ$
to $13.1^\circ$ on RICH, and increases MOOF N-FKE\textsubscript{2d} from $0.67$
to $0.91$. We hypothesize that knee-relative ankle rotations occupy a narrower
training distribution, whereas global rotations better support generalization
to the complex foot poses in MOYO and MOOF~\cite{wehrbein26footmr}.

\subsubsection{Global-Orientation Augmentation}
Removing the random global-orientation augmentation increases MOYO and RICH
AJAE to $22.9^\circ$ and $13.4^\circ$, respectively, and MOOF
N-FKE\textsubscript{2d} to $0.81$. Because FootMR does not use image features,
this augmentation can consistently transform its 2D and rotation inputs and
expose FootMR to a broader range of global ankle orientations.

\newpage
\subsection{Additional Qualitative Results}
\label{sec:supplement_qualitative}
\begin{figure*}[t]
    \centering
    \includegraphics{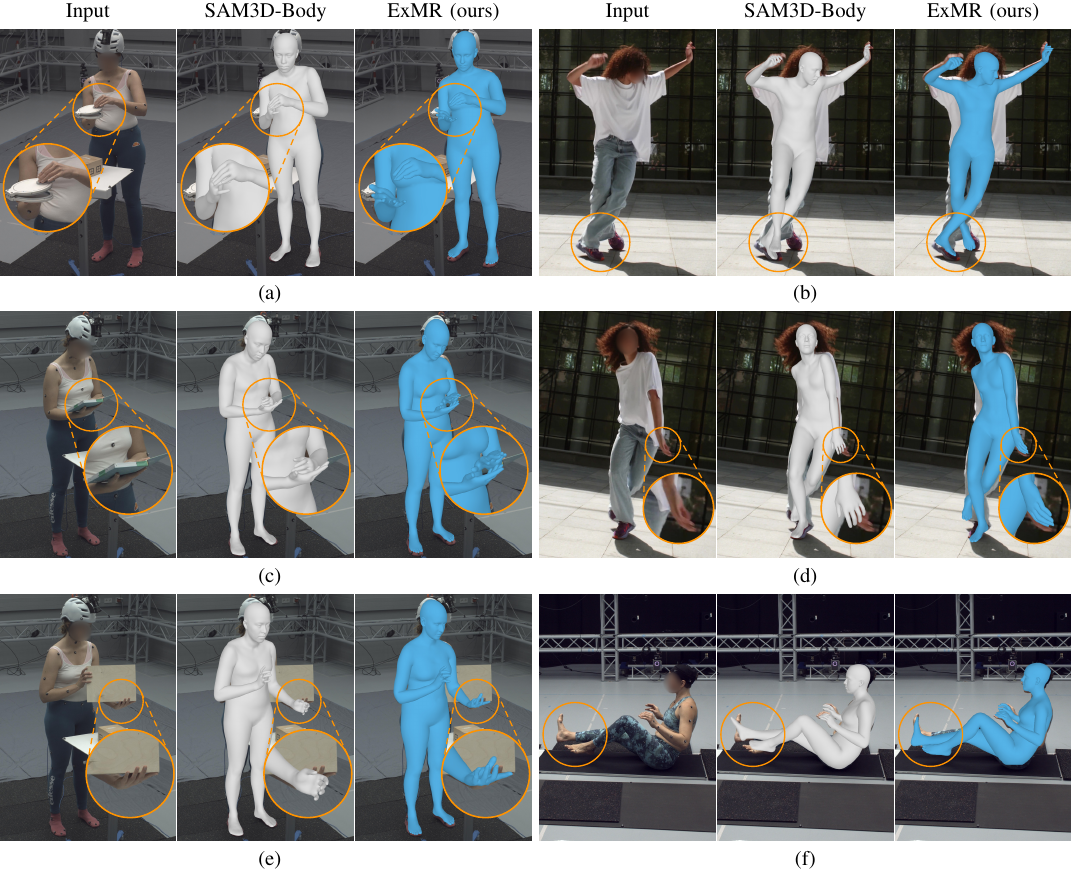}\par\prevdepth=207.68831pt\relax
    \caption{\textbf{Qualitative comparison with SAM3D-Body.}
        Each example shows, from left to right, the input image, the
        SAM3D-Body prediction (gray), and the \modelname\ prediction (blue).
        Panels (a), (c), and (e) show object-manipulation examples;
        (b) and (d) show dance poses; (f) shows a seated pose.
        Within each example, corresponding insets use identical image regions
        and magnification.
        SAM3D-Body overlays show its native MHR meshes, while
        \modelname\ overlays show SMPL-X meshes.
    }\label{fig:sam3d_comparison}
\end{figure*}

Fig.~\ref{fig:sam3d_comparison} compares \modelname\ with
SAM3D-Body~\cite{yang2026sam3dbody} on six
examples, showing the input image alongside both predicted mesh overlays.
These examples include partial hand occlusion by manipulated objects,
pronounced wrist and ankle articulation, and motion blur, which make the
extremity poses difficult to estimate. The highlighted regions show
that \modelname's predictions more closely follow the visible hand
configurations and foot orientations than those of
SAM3D-Body, the strongest per-frame method in our
comparison. Unlike SAM3D-Body, which processes each image independently,
\modelname\ combines observations across neighboring frames. This temporal
context may help resolve ambiguities when an extremity is occluded or blurred
in a particular frame, by exploiting evidence from other frames in the sequence.

Still images illustrate pose alignment but cannot convey the continuity of
the reconstructed motion. We therefore refer readers to the supplementary
video (available on the project page, \url{https://twehrbein.github.io/exmr-website/}) to assess full-body motion over
time, including hand and foot articulation, and the temporal consistency of
the reconstructions.

\FloatBarrier

\end{document}